\documentclass[journal]{IEEEtran}

\usepackage{times}
\usepackage{amsmath,amsfonts,amssymb,bm}
\usepackage{array}
\usepackage{graphicx}
\usepackage{url}
\usepackage{xcolor}
\usepackage{soul}
\usepackage{booktabs}
\usepackage{multirow}
\usepackage{cite}
\usepackage[breaklinks=true]{hyperref}
\usepackage{amsthm}

\usepackage{algorithm}
\usepackage{algorithmic}
\usepackage{cite}                  
\usepackage{graphicx}              
\usepackage{amsmath,amssymb}       
\usepackage{booktabs}              
\usepackage{hyperref}  
\usepackage{algorithm}
\usepackage{algorithmic}

\ifCLASSINFOpdf
\else
\fi
\begin{document}
%
\title{Additive Causal Construction for Transferable and Reconfigurable Cross-System Learning in Multi-Source Image Fusion}
%
%
%

\author{
Zhizhong~Fu$^{1,*}$,
Wei~Zhou$^{1,5,7,*}$,
Zhaoyang~Jiang$^{2}$,
Yudong~Lin$^{1}$,
Yifu~Hou$^{3}$,
Yuke Cao$^{4}$,
Xiaorong~Ding$^{1}$,
Qiang~Yan$^{6}$,
and~Yifan~Chen$^{1}$%
\thanks{$^{*}$Zhizhong Fu and Wei Zhou contributed equally to this work.}
\thanks{Corresponding author: Yifan Chen (yifan.chen@uestc.edu.cn).}
\thanks{$^{1}$School of Life Science and Technology, University of Electronic Science and Technology of China, Chengdu 610054, China.}
\thanks{$^{2}$School of Health and Wellbeing, University of Glasgow, Glasgow G12 8TB, United Kingdom.}
\thanks{$^{3}$Department of Organ Transplantation, Sichuan Provincial People’s Hospital, University of Electronic Science and Technology of China, Chengdu, China.}

\thanks{$^{4}$School of Information and Software Engineering, University of Electronic Science and Technology of China, Chengdu 611731, China.}
\thanks{$^{5}$Department of Radiology, Huzhou Maternity \& Child Health Care Hospital, Huzhou 313000, China.}
\thanks{$^{6}$Hepatological Surgery Department, Huzhou Central Hospital, Fifth School of Clinical Medicine of Zhejiang Chinese Medical University, Huzhou 313000, China.}
\thanks{$^{7}$Department of Radiology, Huzhou Maternity \& Child Health Care Hospital Affiliated to Huzhou Normal University,Huzhou 313000,China.}

}

%
%

\markboth{Journal of \LaTeX\ Class Files,~Vol.~14, No.~8, August~2015}%
{Shell \MakeLowercase{\textit{et al.}}: Bare Demo of IEEEtran.cls for IEEE Journals}
%



\maketitle
\begin{abstract}
In multi-source image fusion scenarios, heterogeneous inputs are typically driven by distinct generative mechanisms and can be viewed as a composition of multiple causal systems. However, cross-system discrepancy (CSD) and cross-system entanglement (CSE) commonly arise during the fusion process, often leading to significant performance degradation under out-of-distribution (OOD) predictions.
To address the CSD and CSE issues, we propose the additive causal construction (ACC) framework, which characterizes information fusion at two levels: firstly, it establishes causal ``anchors'' shared among multiple systems through intervention consistency to enable causal graph transferability (CGT); and secondly, it formalizes the fusion process as causal construction and models the reliability of constructed paths through uncertainty quantification to ensure causal graph reconfigurability (CGR).
Building upon this, we revisit the traditional causal representation learning (CRL) with ACC and propose ACC-CRL as a learnable instantiation of the framework. The method explores joint causal content representations across systems \textit{via} content–mechanism decoupling, and performs response alignment under shared anchors to mitigate CSD. Furthermore, it incorporates structural uncertainty to adaptively regulate the fusion process, thereby suppressing unstable CSE.
We conduct systematic experiments on synthetic data (ColorMNIST) and real-world multi-center medical imaging tasks (microvascular invasion (MVI) prediction). The results demonstrate that the proposed method significantly improves OOD generalization while maintaining in-distribution (ID) performance, validating the effectiveness and robustness of the ACC-CRL strategy based on mechanism alignment and uncertainty modeling in open environments.\end{abstract}



%
\IEEEpeerreviewmaketitle

\section{Introduction}
Multi-source data are often generated under distinct principles and can be formalized as multiple causal systems with heterogeneous mechanisms. In this context, the central challenge of image fusion extends beyond simple information aggregation to addressing heterogeneity in mechanisms. Firstly, the absence of a unified reference for causal mechanisms across systems makes semantic alignment difficult, leading to the cross-system discrepancy (CSD) issue. Secondly, the fusion process may introduce unstable dependencies on cross-system interference, leading to spurious causal pathways that are not robust to distribution shifts, a phenomenon referred to as cross-system entanglement (CSE).

These two challenges are particularly evident in medical image analysis. Mechanism heterogeneity arises not only from variations in scanning devices, acquisition protocols, and population distributions across centers, but also from differences in biological processes across spatial regions within a single image. For instance, in tumor analysis, intratumoral and peritumoral regions often correspond to distinct pathological mechanisms and generative processes. Therefore, multi-source fusion should be treated as a cross-system causal inference problem rather than a simple multi-input modeling task. However, existing approaches typically assume direct combinability of inputs from different sources, without explicitly modeling differences in generative mechanisms or the resulting changes in causal structures after fusion, leading to significant performance degradation under distribution shifts.

Prior studies have explored anomaly detection, robust representation learning, and causal intervention to address distribution shifts. For example, MultiOOD~\cite{dong2024multiood} utilizes prediction discrepancies across multiple branches for out-of-distribution (OOD) detection; IHF~\cite{vasiliuk2023limitations} improves robustness through statistical feature modeling; and SI-CRL~\cite{LIU2025103741} learns invariant representations \textit{via} frequency-domain interventions. Although these methods alleviate distribution shifts to some extent, they are largely grounded in an observational-level, joint modeling paradigm, in which the causal structure is treated as a pre-specified, static object, and robustness is achieved \textit{via} static constraints. As a result, they fall short of capturing the more fundamental problems of CSD and CSE. 

To overcome the aforementioned issues, in this work, we do not analyze the causal graph from the \textit{a posteriori} perspective, where its structure is fixed. Instead, we view the graph from the \textit{a priori} perspective, where its topology is changeable, emerging from meticulously designed interactions among heterogeneous mechanisms during multi-source fusion. In this way, we describe the fusion process as a dynamic generation of causal structures across systems using model priors, enabling adaptive regulation of causal relationships based on mechanism compatibility and structural uncertainty.

Subsequently, we propose the additive causal construction (ACC) framework, which models each input source as a causal system with distinct generative mechanisms and formulates information fusion as a progressive process of constructing a unified causal structure, \textcolor{black}{where the term ``additive'' refers to the integration of multiple separated causal systems into a unified structure through progressively ``adding'' complementary cross-system causal relations. }Specifically, the framework is developed from the following two key aspects: (i) causal graph transferability (CGT), where shared causal ``anchors'' across multiple systems are established through intervention consistency to mitigate CSD; and (ii) causal graph reconfigurability (CGR), where fusion is treated as topological reorganization, with uncertainty modeling used to evaluate and regulate candidate causal pathways, thereby suppressing unstable CSE. Building upon ACC, we further revisit the classical causal representation learning (CRL) and introduce the ACC-CRL implementation. The method learns shared causal content representations across systems through content--mechanism decoupling and performs response alignment under shared anchors to mitigate CSD. Meanwhile, structural uncertainty is incorporated to enable adaptive fusion, suppress CSE, and approximate construction of causal diagrams in the representation space.

Furthermore, we conduct comprehensive experiments on both synthetic data (ColorMNIST) and real-world multi-center medical imaging tasks (microvascular invasion (MVI) prediction). In the MVI task, a single-sequence dual-region input is employed to simulate cross-region fusion under heterogeneous mechanisms. The results show that the proposed method significantly improves OOD generalization while maintaining in-distribution (ID) performance, demonstrating the effectiveness of the proposed ACC-CRL strategy based on mechanism alignment and uncertainty modeling.

The remainder of this paper is organized as follows. Section II reviews related works on multi-source fusion under distribution shifts and MVI prediction. Section III introduces the proposed ACC framework, providing a unified formulation of cross-system causal construction in terms of CGT and CGR. Section IV presents ACC-CRL as a learnable instantiation of the framework, detailing its representation learning, mechanism alignment, and uncertainty-aware fusion strategies. Section V reports extensive experiments on both synthetic and real-world medical imaging datasets to validate the effectiveness and robustness of the proposed method. Finally, Section VI concludes the paper and discusses future directions.

\section{Related Works}
\subsection{Multi-Source Fusion under OOD Settings}

Achieving robust generalization under OOD scenarios is a fundamental challenge in machine learning. Traditional methods based on empirical risk minimization tend to overfit spurious correlations in the training distribution, leading to significant performance degradation under distribution shifts. To address this issue, prior studies have explored data augmentation (e.g., Mixup~\cite{zhang2017mixup}), invariance learning (e.g., IRM~\cite{arjovsky2019invariant}), and distributionally robust optimization (DRO~\cite{sagawa2019distributionally}) to improve cross-distribution stability.

However, these approaches are primarily designed for single-input settings. In multi-source fusion, the OOD problem becomes more challenging: different input sources may provide complementary information, but their joint modeling can also introduce spurious dependencies, resulting in unstable cross-source correlations. Recent studies have begun to address this issue. For example, MultiOOD~\cite{dong2024multiood} detects OOD samples by leveraging prediction discrepancies across multiple input branches; CRL methods (e.g., MMNAR~\cite{liang2025causal}) further consider the non-random missingness of modalities; and contrastive learning approaches (e.g., SupContrast~\cite{khosla2020supervised}) enhance robustness by improving representation discriminability.

While the above approaches alleviate distribution shifts to some extent, they largely rely on assumptions of statistical consistency and improve robustness through static constraints at the observational level. As a result, they lack explicit modeling of differences in generative mechanisms and the evolution of causal structures, making them inadequate for handling unstable CSE pathways introduced by cross-source interference. Fundamentally, such methods assume that the causal structure is pre-existing and fixed, imposing constraints upon it rather than modeling its generation and evolution. In contrast, this work adopts a causal construction perspective, framing CSE as a process of dynamic generation of causal structures, and achieves adaptive regulation of CSE through mechanism alignment and structural uncertainty.

\subsection{MVI Prediction}
MVI is a critical risk factor for postoperative recurrence and poor prognosis in hepatocellular carcinoma (HCC), and its accurate preoperative prediction is of great clinical importance~\cite{li2023preoperative,zhang2024deep}. Early approaches primarily relied on single clinical or imaging features, such as serum biomarkers and radiological characteristics, but suffered from limited predictive performance and stability~\cite{wang2022mvi}. Subsequently, radiomics-based methods improved performance by extracting high-dimensional handcrafted features combined with machine learning models, yet their generalization ability and reproducibility remain limited~\cite{jiang2022predicting}.

With the advancement of deep learning, end-to-end models have become dominant. For instance, three-dimensional convolutional neural network-based methods can extract spatial and semantic information from multi-sequence MRI, while approaches incorporating Transformers or topological modeling further enhance feature representation and offer partial interpretability~\cite{zheng2025mri,zhang2021deep}. In addition, studies integrating imaging, clinical, and pathological information have demonstrated improved MVI prediction performance~\cite{wang2024mri}.

Existing methods still face limited generalization under OOD settings. First,  most approaches rely on statistical learning paradigms without explicitly modeling the underlying causal structures and their uncertainties, overlooking CSD. Second, heterogeneous mechanisms across input sources often lead to unstable or spurious dependencies, reflecting CSE. In this work, the MVI task involves intra-tumoral and peri-tumoral regions from the same arterial-phase (ART) MRI sequence, making it a cross-region fusion problem rather than a strictly multi-modal one. Despite this, the regions provide complementary information and may exhibit region-specific mechanisms under multi-center settings, forming a meaningful image fusion challenge.

To address this, we approach the problem through ACC, dynamically regulating the fusion process by modeling CGT and CGR, thus improving cross-distribution generalization in MVI prediction. Unlike existing methods, our approach explicitly tackles CSD and CSE from a causal perspective.

\section{ACC Framework}

\begin{figure*}[!t]
    \centering
    \includegraphics[width=1\textwidth]{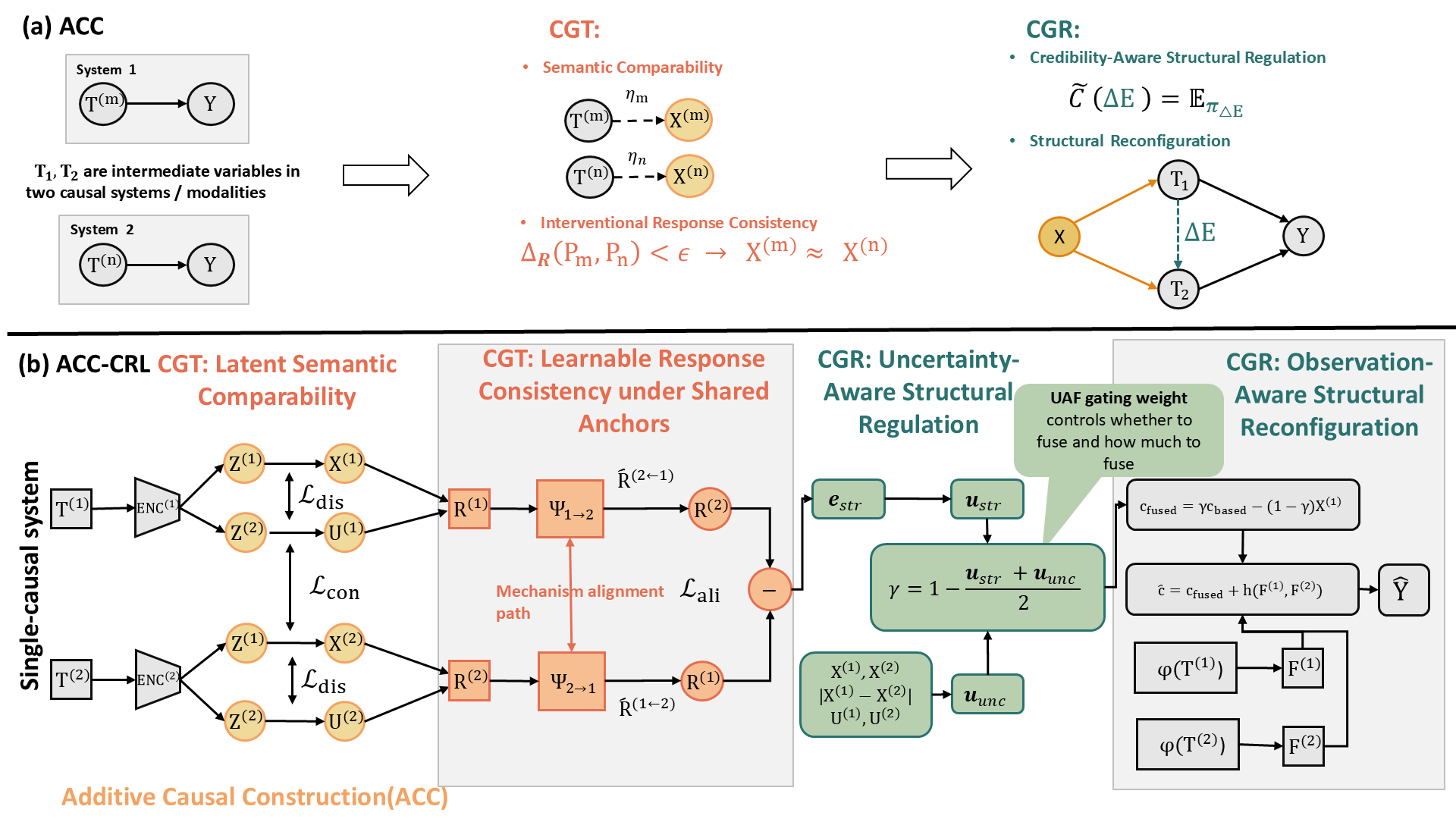}
    \caption{
Overview of the proposed ACC framework and its trainable realization, ACC-CRL.
(a) ACC formulates multi-source fusion as a cross-system causal construction process under heterogeneous mechanisms. First, CGT establishes shared causal anchors across systems through semantic comparability and interventional response consistency, enabling comparable causal semantics between heterogeneous systems. Second, CGR dynamically reconstructs cross-system causal relations through credibility-aware structural regulation and structural reconfiguration, where candidate cross-system edges $\Delta\mathbf E$ are selectively preserved or suppressed according to structural credibility.
(b) ACC-CRL provides a learnable realization of ACC in representation space. The CGT module learns latent semantic comparability through content–mechanism disentanglement and further diagnoses cross-system structural consistency \textit{via} bidirectional response alignment under shared anchors. The CGR module subsequently uses the resulting structural-mismatch signal for uncertainty-aware structural regulation, in which structural and predictive uncertainties jointly determine the reliability of adaptive fusion. Finally, observation-aware structural reconfiguration supplements latent causal representations with observation-level visual features to improve robustness under distribution shifts.
}
    \label{fig:general plan}
\end{figure*}

As shown in Fig.~\ref{fig:general plan}(a), ACC treats multi-source data as originating from heterogeneous causal systems and formulates fusion as a dynamic process of causal construction rather than simple feature aggregation.

This formulation is motivated by two coupled challenges. First, due to heterogeneous generative mechanisms, different systems lack a unified causal reference, making their interventional effects difficult to compare directly. This gives rise to CSD. To mitigate CSD, ACC introduces CGT, which establishes shared causal anchors across systems and enforces interventional consistency, thereby providing a common causal alignment basis.
Second, the fusion process may reorganize existing causal pathways and introduce new cross-system links. When such links are induced by mechanism mismatches or unreliable correlations between systems, they may form unstable dependencies, leading to CSE. To monitor CSE, ACC introduces CGR, which treats fusion as the reconstruction of the underlying causal diagram and uses structural uncertainty to evaluate and regulate candidate causal pathways.

\subsection{Principles of CGT}

Let the $m^\mathrm{th}$ and $n^\mathrm{th}$ causal systems be defined as
\begin{align}
\mathcal G^{(m)}=\big\langle \mathbf V^{(m)},\mathbf E^{(m)},\mathbf U^{(m)}\big\rangle,~~~
\mathcal G^{(n)}=\big\langle \mathbf V^{(n)},\mathbf E^{(n)},\mathbf U^{(n)}\big\rangle,
\end{align}
where $\mathbf V^{(\cdot)}$ is the node set, $\mathbf E^{(\cdot)}$ is the directed causal edge set, and $\mathbf U^{(\cdot)}$ is the mechanism variables of the corresponding causal system. Since heterogeneous mechanisms govern different systems, their interventional effects are generally not comparable in the original space. Therefore, establishing CGT is a prerequisite for multi-source fusion.

Let $v_i^{(m)}\in \mathbf V^{(m)}$ and $v_j^{(n)}\in \mathbf V^{(n)}$ denote candidate nodes from two causal systems, respectively. 
Subsequently, a \textit{shared anchor} $X$ is defined as an abstract causal content
reference induced by cross-system alignment. It does not require the existence
of an identical raw variable in different systems. Correspondingly,
an \textit{anchor-equivalent pair} refers to two system-specific nodes
that can be associated with the same shared anchor and induce consistent
interventional responses.

Next, we determine whether $v_i^{(m)}$ and $v_j^{(n)}$ constitute an
anchor-equivalent pair with respect to the shared anchor $X$ from two
perspectives: semantic comparability and interventional response consistency.

\subsubsection{Semantic Comparability}
Prior work on cross-model causal consistency has shown that variables from different causal models need not be identical in their original spaces; rather, a causal correspondence can be established as long as suitable transformations yield consistent causal predictions under corresponding interventions \cite{rubenstein2017causal}. Accordingly, semantic comparability requires that candidate nodes from different systems can be mapped into a common causal content space.

Formally, suppose there exist measurable alignment mappings
\begin{align}
\eta_m:\mathcal D_i^{(m)}\rightarrow \mathcal D_X,\qquad
\eta_n:\mathcal D_j^{(n)}\rightarrow \mathcal D_X,
\end{align}
such that
\begin{align}
X^{(m)}=\eta_m\left(v_i^{(m)}\right),\qquad
X^{(n)}=\eta_n\left(v_j^{(n)}\right).
\end{align}
Then $v_i^{(m)}$ and $v_j^{(n)}$ are said to be comparable in the shared causal content space $\mathcal D_X$. Here, $\mathcal D_i^{(m)}$ and $\mathcal D_j^{(n)}$ are the value spaces of the candidate nodes in their respective systems, and $\mathcal D_X$ is the value space of $X$. The variables $X^{(m)}$ and $X^{(n)}$ denote the system-specific anchor representations induced in the shared causal content space \textcolor{black}{\textit{via} the CGT process shown in Fig.~\ref{fig:general plan}(a).}

Importantly, semantic comparability only establishes a common reference space; it does not imply that $X^{(m)}$ and $X^{(n)}$ already correspond to the same causal node. Their identification as a common abstract anchor further requires the following interventional response consistency.

\subsubsection{Interventional Response Consistency}
Invariant causal prediction and risk minimization posit that stable causal relations should remain invariant across environments or mechanism shifts \cite{peters2016causal,arjovsky2019invariant}. Therefore, once a shared causal content reference is established, the two system-specific anchor representations need not share identical generative mechanisms, but only consistent causal semantics:
\begin{align}
X^{(m)} \approx X^{(n)},
\label{eq:invariant_content_consistency}
\end{align}
where the approximate consistency indicates correspondence in shared causal content, rather than duplication of the same node in the original causal graphs. \textcolor{black}{As illustrated by the CGT module in Fig.~\ref{fig:general plan}(a), $X^{(m)}$ and $X^{(n)}$ are mapped into a shared causal content space and treated as anchor-equivalent representations when they induce consistent interventional responses.}

Furthermore, under the corresponding interventions on the shared causal content, the induced downstream responses in the two systems should also remain consistent:
\begin{align}
\Delta_R\left(
P_m\left(R_X^{(m)}\mid do(X^{(m)}=x)\right),
P_n\left(R_X^{(n)}\mid do(X^{(n)}=x)\right)
\right)\nonumber\\
\leq \epsilon_X,\quad \forall x\in\mathcal D_X .
\label{eq:anchor_response_consistency}
\end{align}
where $\Delta_R(\cdot,\cdot)$ denotes a response-level distribution discrepancy function, which measures the mismatch between the two interventional response distributions induced by the same shared
anchor intervention. The causal effect of anchor $X$ refers to the change in the downstream response induced by an intervention on the anchor representation. In system $m$, this effect is characterized by the interventional response distribution
$P_m\left(R_X^{(m)}\mid do(X^{(m)}=x)\right)$, which describes how the downstream response
$R_X^{(m)}$ would behave if the system-specific anchor representation
$X^{(m)}$ is externally set to $x$. The term $\epsilon_X$ is the tolerance threshold for approximate causal equivalence. The special case $\epsilon_X=0$ reduces to strict interventional consistency. The variables $R_X^{(m)}$ and $R_X^{(n)}$ denote the corresponding descendant responses induced by $X^{(m)}$ and $X^{(n)}$ in the two systems, respectively.

Finally, if $X^{(m)}$ and $X^{(n)}$ satisfy both semantic comparability and interventional response consistency, then $v_i^{(m)}$ and $v_j^{(n)}$ are anchor-equivalent with respect to $X$, denoted by
\begin{align}
v_i^{(m)} \overset{\mathrm X}{\sim} v_j^{(n)},
\end{align}
where $X$ should be understood as the representative of the corresponding equivalence class, rather than as a direct merger of the original nodes. Hence, the role of CGT is not to identify perfectly identical variables across systems, but to construct cross-system shared anchors that provide a comparable causal reference for subsequent CGR, thereby enabling ACC to further regulate candidate causal relations under heterogeneous mechanisms. The theoretical connection between interventional response consistency and first-order mechanism alignment is further analyzed in Appendix \ref{appendix_a}.

\subsection{Principles of CGR}
\subsubsection{Credibility-Aware Structural Regulation}
\textcolor{black}{Since candidate relations constructed across heterogeneous systems may exhibit different levels of structural stability and reliability, CGR introduces a unified structural uncertainty formulation to evaluate candidate causal edges before reconstruction.} For any candidate edge $e=(v_i\rightarrow v_j)$, let
\begin{align}
\pi_e\sim P(\pi_e),
\end{align}
where $\pi_e$ represents structural priors, mechanism compatibility, or other
sources of uncertainty associated with the edge, and $P(\pi_e)$ denotes the
prior distribution over such structural confidence variables. The expected credibility of the edge is then defined as
\begin{align}
\tilde C(e)
=
\mathbb E_{\pi_e}
\Big[
\mathbb E\big[C(e)\mid \pi_e\big]
\Big],
\end{align}
where $C(e)$ denotes the causal credibility of the edge under a given structural confidence condition. This credibility score determines whether a candidate edge should be preserved, replaced, weakened, or suppressed during CGR.

Under this unified credibility measure, CGR handles the two forms of structural variation separately.

First, for conflicting existing causal relations, CGR does not directly merge
edges from different systems. Instead, it performs competitive selection based
on causal credibility. Let
\begin{align}
e^{(m)}_{ij}\in \mathbf E^{(m)},\qquad
e^{(n)}_{ij}\in \mathbf E^{(n)},
\end{align}
where $e^{(m)}_{ij}$ and $e^{(n)}_{ij}$ are candidate causal edges from systems
$m$ and $n$, respectively. The subscript $ij$ indexes the corresponding
source--target structural role, while the superscripts $(m)$ and $(n)$ indicate
the systems from which the edges originate. Since these edges may arise from
different mechanisms, their stability and reliability can differ. CGR therefore
selects the more reliable relation according to
\begin{align}
e_{ij}^{*}
=
\arg\max_{e\in\{e^{(m)}_{ij},e^{(n)}_{ij}\}}
\tilde C(e),
\end{align}
where $e_{ij}^{*}$ corresponds to the selected causal relation for the structural
role indexed by $ij$ after credibility-based competition.
This process characterizes the preservation, replacement, or suppression of
existing causal relations during fusion.

Second, for newly introduced cross-system edges, let
\begin{align}
\Delta\mathbf E_{ij}
=
\left\{e^{\mathrm{cross}}_{ij,k}\right\}_{k=1}^{K}
\subseteq \Delta\mathbf E
\end{align}
denote the set of candidate cross-system causal edges proposed for the
source--target structural role indexed by $ij$. Unlike existing intra-system
relations, these edges do not originate from any single causal graph, but are
constructed under heterogeneous mechanism interactions and constructive priors.
They may encode complementary cross-system information, but may also introduce
unstable dependencies caused by mechanism mismatch or representational
heterogeneity.

For each candidate cross-system edge $e\in\Delta\mathbf E_{ij}$, its
cross-system credibility is defined as
\begin{align}
\tilde C_{\mathrm{cross}}(e)
=
\mathbb E\big[C(e)\mid \pi_e^{\mathrm{cross}}\big],
\end{align}
where $\pi_e^{\mathrm{cross}}$ is jointly determined by mechanism compatibility,
anchor response consistency, and structural uncertainty. \textcolor{black}{As illustrated by the
CGR process in Fig.~\ref{fig:general plan}(a), these constructive structural
priors are used to evaluate the reliability of candidate cross-system causal
edges before they are preserved, weakened, or suppressed during causal
construction.} CGR then selects the
most reliable cross-system relation according to
\begin{align}
e_{ij}^{\mathrm{cross},*}
=
\arg\max_{e\in\Delta\mathbf E_{ij}}
\tilde C_{\mathrm{cross}}(e),
\end{align}
where $e_{ij}^{\mathrm{cross},*}$ denotes the selected cross-system causal
relation for the structural role indexed by $ij$. This edge is retained only when its credibility exceeds
a reliability threshold $\delta$, i.e.,
\begin{align}
\tilde C_{\mathrm{cross}}\left(e_{ij}^{\mathrm{cross},*}\right) \geq \delta .
\end{align}
Otherwise, the corresponding relation is weakened or suppressed during
reconstruction. Large discrepancies in cross-system responses indicate that
the candidate edge may correspond to an unstable shortcut dependency, whereas
consistent responses suggest that the edge is more likely to encode stable
causal semantics.
\subsubsection{Structural Reconfiguration}
\textcolor{black}{With the above credibility-aware regulation principle, CGR further formulates
fusion as a structural reconfiguration process. After shared anchors are
established across systems, fusion is no longer treated as a simple
concatenation or extension of pre-existing causal graphs. Instead, CGR
constructs a candidate causal system by reorganizing existing intra-system
relations and introducing credibility-regulated cross-system relations under
the guidance of model priors. In this view, the causal diagram is not
passively recovered as a fixed structure, but progressively constructed and
regulated under heterogeneous mechanisms through shared anchors, edge-level
credibility, and structural uncertainty.}

Specifically, joint modeling across multiple causal systems introduces two types of structural variations: (i) the competitive selection among existing intra-system causal relations, and (ii) the emergence of novel cross-system candidate causal paths. The former corresponds to the preservation, replacement, or suppression of existing relations during fusion, while the latter reflects structural expansion induced by interactions between heterogeneous mechanisms. Since these candidate structures may contain both stable causal relations and unstable shortcut dependencies, structural uncertainty is introduced to quantify and regulate their reliability.

Accordingly, the candidate causal system constructed during fusion is defined as
\begin{align}
\mathcal G'
=
\langle \mathbf V',\mathbf E',\mathbf U'\rangle,
\end{align}
where the node set is given by
\begin{align}
\mathbf V'
=
\mathbf V^{(m)}\cup \mathbf V^{(n)}\cup \mathbf V_X .
\end{align}
Here, $\mathbf V_X$ denotes the set of shared causal content nodes induced by
anchor-equivalence relations. Specifically, if $X^{(m)}$ and $X^{(n)}$ satisfy
semantic comparability and interventional response consistency, they are
identified as the same abstract anchor $X\in\mathbf V_X$ at the level of
shared causal content.

The fused candidate edge set satisfies
\begin{align}
\mathbf E'
\subseteq
\mathbf E^{(m)}\cup \mathbf E^{(n)}\cup \mathbf E_X\cup \Delta\mathbf E ,
\end{align}
where $\mathbf E_X$ denotes constructive alignment relations induced by
anchor equivalence, which provide shared references for cross-system
structural regulation, and \textcolor{black}{$\Delta\mathbf E$ denotes the newly introduced cross-system candidate causal edges generated during the CGR process in Fig.~\ref{fig:general plan}(a). }Importantly, $\mathbf E_X$ should not
be interpreted as direct copies or mergers of original causal edges; rather,
it represents auxiliary alignment constraints introduced by shared anchors
during reconstruction.

Finally, prediction is performed on the constructed candidate causal system $\mathcal G'$:
\begin{align}
\hat Y
=
f_Y(\mathbf X;\mathcal G'),
\end{align}
where $f_Y(\cdot;\mathcal G')$ is the prediction function defined on the
constructed candidate causal system $\mathcal G'$, and $\hat Y$ is the
predicted outcome.
This process reflects the core principle of CGR: under heterogeneous mechanisms, reliable causal diagrams are constructed through prior-guided regulations rather than naive concatenation, thereby enabling robust cross-system reasoning.

\section{ACC-CRL as Trainable Realization of ACC}

ACC formulates multi-source fusion as a cross-system causal construction process regulated by model priors. However, ACC itself remains a conceptual formulation and must be instantiated as a learnable representation-space mechanism. To this end, we propose ACC-CRL as an end-to-end trainable realization of ACC.

The core idea of ACC-CRL is threefold. First, observations from each system are decomposed in latent space into shared causal content and system-specific mechanisms, thereby constructing comparable anchors across systems. Second, bidirectional response alignment is performed under shared anchors, transforming cross-system response inconsistency into explicit structural mismatch signals. Third, structural mismatch and predictive uncertainty are jointly used to construct sample-level structural regulation variables that adaptively control cross-system fusion strength. Consequently, ACC-CRL does not perform unconstrained multi-source aggregation, but instead approximates the causal construction process in representation space under the principle of \textit{constructing shared anchors, diagnosing structural mismatch, fusing when reliable, and suppressing when unreliable.}

\subsection{Learnable Realization of CGT}
\subsubsection{Latent Semantic Comparability}
Let $T^{(1)}$ and $T^{(2)}$ denote observations from two causal systems. To realize the shared-anchor mechanism in CGT, ACC-CRL encodes each observation into a shared causal content variable and a system-specific mechanism variable:
\begin{equation}
\bigl(X^{(m)}, U^{(m)}\bigr)
=
\mathcal{C}_m\!\left(T^{(m)}\right),
\quad m\in\{1,2\}.
\label{eq:ACC_crl_encoder}
\end{equation}
where $T^{(m)}$ is the observation from system $m$, and
$\mathcal{C}_m$ is the corresponding system-specific encoder.
The term $X^{(m)}$ denotes the shared causal content representation induced from
system $m$, corresponding to the system-specific anchor representation
introduced previously, and $U^{(m)}$ captures system-specific mechanism
factors that characterize generative heterogeneity across systems. This
decomposition reflects the CGT principle that shared causal content nodes
are induced by anchor-equivalent relations rather than direct mergers of
original variables.

To ensure comparability between content representations across systems, we impose a contrastive consistency constraint on paired samples:
\begin{equation}
\mathcal{L}_{\mathrm{con}}
=
-\sum_i
\log
\frac{
\exp\!\left(\mathrm{sim}\left(X_i^{(1)},X_i^{(2)}\right)/\tau\right)
}{
\sum_j
\exp\!\left(\mathrm{sim}\left(X_i^{(1)},X_j^{(2)}\right)/\tau\right)
},
\label{eq:ACC_crl_con}
\end{equation}
where $\mathrm{sim}(\cdot,\cdot)$ is a similarity function and $\tau$ is a temperature parameter. This objective encourages semantically corresponding samples from different causal systems to align in a shared content space, thereby providing comparable anchor references for subsequent response alignment and structural regulation. The two indices $i$ and $j$ indicate a paired sample across the two systems and
candidate samples in the contrastive set, respectively.

Furthermore, to prevent system-specific mechanism information from leaking into the shared content variable, we introduce a content--mechanism disentanglement constraint:
\begin{equation}
\mathcal{L}_{\mathrm{dis}}
=
\sum_{m\in\{1,2\}}
\mathcal{H}\left(X^{(m)},U^{(m)}\right),
\label{eq:ACC_crl_dis}
\end{equation}
where $\mathcal{H}(\cdot,\cdot)$ is instantiated using the Hilbert-Schmidt
Independence Criterion (HSIC) with Gaussian kernels. This constraint reduces redundant coupling between $X^{(m)}$ and $U^{(m)}$, encouraging $X^{(m)}$ to encode stable cross-system causal content while preserving mechanism-specific variation in $U^{(m)}$.

Accordingly, anchor construction is transformed in ACC-CRL into a latent-space process of shared content learning with mechanism disentanglement, \textcolor{black}{as illustrated in the CGT realization of Fig.~\ref{fig:general plan}(b). $X^{(1)}$ and $X^{(2)}$ provide comparable cross-system causal content references, while $U^{(1)}$ and $U^{(2)}$ explicitly preserve mechanism heterogeneity.}

\subsubsection{Learnable Response Consistency under Shared Anchors}

After establishing shared anchor references through content--mechanism disentanglement, ACC-CRL further diagnoses whether cross-system structures remain consistent under the shared reference. Specifically, for system $m\in\{1,2\}$, define its task response conditioned on content and mechanism as
\begin{equation}
R^{(m)}
=
f_Y^{(m)}\!\left(X^{(m)},U^{(m)}\right),
\label{eq:response}
\end{equation}
where $R^{(m)}$ denotes the task-head output, e.g., logits, representing the system response under the current anchor content and mechanism condition.

To characterize whether the mechanism of one system can explain the response of another under the shared anchor reference, we construct bidirectional cross-system response mappings:
\begin{equation}
\begin{aligned}
\hat{R}^{(2\leftarrow 1)}
&=
\Psi_{1\rightarrow 2}\!\left(X^{(1)},U^{(1)},U^{(2)}\right),\\
\hat{R}^{(1\leftarrow 2)}
&=
\Psi_{2\rightarrow 1}\!\left(X^{(2)},U^{(2)},U^{(1)}\right),
\end{aligned}
\label{eq:cross_mapping}
\end{equation}
where $\Psi_{1\rightarrow 2}$ and $\Psi_{2\rightarrow 1}$ are learnable response mappings. Rather than explicitly defining theoretical mechanism transformation operators, they reconstruct one system's response using the content representation from one side, together with the mechanism representations from both sides. Thus, $\Psi$ serves as a learnable approximation of the cross-system mechanism alignment principle in CGT.

If two systems share consistent causal semantics under the same anchor, then the mapped response from one system should explain the response of the other. Accordingly, the bidirectional response alignment residuals are defined as
\begin{equation}
e_{12}
=
\left\|
\hat{R}^{(2\leftarrow 1)}-R^{(2)}
\right\|_2^2,
\quad
e_{21}
=
\left\|
\hat{R}^{(1\leftarrow 2)}-R^{(1)}
\right\|_2^2.
\label{eq:residual}
\end{equation}
The corresponding response alignment objective is
\begin{equation}
\mathcal{L}_{\mathrm{ali}}
=
\sum_i
\left(
\left\|
\hat{R}^{(2\leftarrow 1)}_i-R^{(2)}_i
\right\|_2^2
+
\left\|
\hat{R}^{(1\leftarrow 2)}_i-R^{(1)}_i
\right\|_2^2
\right).
\label{eq:align_loss}
\end{equation}

\textcolor{black}{As illustrated by the model structure in Fig.~\ref{fig:general plan}(b),} $\mathcal{L}_{\mathrm{ali}}$ is not merely a feature-alignment objective. A theoretical discussion on how the proposed interventional response alignment suppresses spurious cross-system mechanism correlations is provided in Appendix~\ref{sec:spurious_correlation}. In ACC-CRL, it additionally serves as a structural diagnosis mechanism: if two systems cannot mutually explain responses under the shared anchor reference, then their underlying causal effects are inconsistent, suggesting that the corresponding cross-system relations may be unreliable. Accordingly, we define the average structural error
\begin{equation}
e_{\mathrm{str}}
=
\frac{1}{2}
\left(
e_{12}+e_{21}
\right),
\label{eq:struct_error}
\end{equation}
as a sample-level structural mismatch signal. This signal acts as a proxy for the reliability of candidate cross-system edges in CGT and is subsequently used for uncertainty modeling and fusion regulation.

\subsection{Learnable Realization of CGR}
\subsubsection{Uncertainty-Aware Structural Regulation}

For CGR, the preservation, weakening, or suppression of candidate cross-system paths is jointly regulated by edge-level credibility and structural uncertainty. In ACC-CRL, this principle is realized through a sample-level uncertainty-aware fusion mechanism.

First, structural uncertainty is defined as:
\begin{equation}
u_{\mathrm{str}}
=
\sigma
\left(
\frac{e_{\mathrm{str}}}{\tau}
\right),
\label{eq:ustruct}
\end{equation}
where $e_{\mathrm{str}}$ is the bidirectional response-alignment
residual defined in \eqref{eq:struct_error}, measuring the degree of
cross-system response inconsistency under the shared anchor; $\sigma(\cdot)$ is the Sigmoid function; and $\tau$ is a temperature parameter.

Beyond structural mismatch, the model must also estimate predictive unreliability at the sample level. To this end, we construct a lightweight uncertainty estimator $g_{\mathrm{unc}}(\cdot)$ with input
\begin{equation}
\xi
=
\left[
U^{(1)},\,
U^{(2)},\,
\left|X^{(1)}-X^{(2)}\right|
\right],
\label{eq:unc_input}
\end{equation}
where $\xi$ is the uncertainty-estimation input, and $|\cdot|$ is the
element-wise absolute difference. The term $g_{\mathrm{unc}}(\cdot)$ maps the mechanism
representations and content discrepancy of the two systems to a scalar score,
which is normalized to obtain the model uncertainty $u_{\mathrm{unc}}$ as
\begin{equation}
u_{\mathrm{unc}}
=
\sigma
\left(
\frac{g_{\mathrm{unc}}(\xi)}{\tau}
\right),
\label{eq:umodel}
\end{equation}
which reflects the predictive risk, whereas $u_{\mathrm{str}}$ in \eqref{eq:ustruct} captures structural mismatch. Together, they regulate fusion from the perspectives of predictive reliability and structural reliability.

Combining both uncertainties yields the total uncertainty from which the cross-system structural regulation variable is defined as
\begin{equation}
\gamma
= 1-
\frac{1}{2}
\left(
u_{\mathrm{unc}}+u_{\mathrm{str}}
\right).
\label{eq:u}
\end{equation}
A larger $\gamma$ indicates stronger structural consistency and more reliable prediction for the current sample, whereas a smaller $\gamma$ suggests that cross-system fusion may introduce unstable dependencies caused by mechanism mismatch.

Let the base fusion representation be
\begin{equation}
c_{\mathrm{base}}
=
\frac{1}{2}
\left(
X^{(1)}+X^{(2)}
\right),
\label{eq:cbase}
\end{equation}
where $c_{\mathrm{base}}$ denotes the naive dual-system content fusion
representation. Here, $X^{(1)}$ is used as the conservative
reference representation when cross-system fusion is unreliable. Then, the uncertainty-regulated causal fusion representation is defined as
\begin{equation}
c_{\mathrm{fused}}
=
\gamma\,c_{\mathrm{base}}
+
(1-\gamma)\,X^{(1)}.
\label{eq:cfused}
\end{equation}
This formulation implies that when cross-system structures are reliable, the model strengthens dual-system fusion; when structural mismatch or predictive risk becomes large, the model suppresses cross-system information and falls back to a more conservative single-system representation.

Accordingly, the abstract notion of cross-system edge credibility in CGR is realized in ACC-CRL as a differentiable sample-level gating mechanism:
\begin{equation}
\tilde C_{\mathrm{cross}}(e)
\propto
\gamma
\label{eq:confidence}
\end{equation}
This mechanism enables ACC-CRL to approximately implement the structural regulation principle of CGR in representation space, preserving and exploiting cross-system relations when reliable, while weakening or suppressing unstable shortcut dependencies when unreliable. A probabilistic interpretation of the adaptive gate $\gamma$ as a sample-wise approximation of cross-system edge credibility is provided in Appendix~\ref{subsec:gamma_derivation}.

\subsubsection{Observation-Aware Structural Reconfiguration}

Although ACC-CRL learns cross-system causal representations through content--mechanism disentanglement in latent space, this abstraction process may weaken certain fine-grained visual cues from the original observations. In medical imaging tasks, boundary morphology, texture patterns, and local structural variations often carry important diagnostic information. Consequently, relying solely on latent causal representations may lead to insufficient utilization of observation-level information.

To address this issue, we introduce an observation-supplementary branch based on raw medical images to preserve discriminative information directly derived from the observation space. Specifically, for each input $T^{(m)}$, an image encoder $\phi(\cdot)$ is constructed to obtain observation feature representations:
\begin{equation}
F^{(m)}
=
\phi\!\left(T^{(m)}\right),
\quad m\in\{1,2\}.
\end{equation}
where $T^{(m)}$ denotes the raw observation from system $m$, and
$\phi(\cdot)$ denotes the image encoder used to extract observation-level
visual features. Here, $F^{(m)}$ primarily captures local visual structures
and serves as a complement to the shared causal content representations
learned in latent space.

Let $c_{\mathrm{fused}}$ denote the causal fusion representation produced by ACC-CRL, and define the aggregated observation representation as
\begin{equation}
c_{\mathrm{obs}}
=
h\!\left(F^{(1)},F^{(2)}\right),
\end{equation}
where $c_{\mathrm{obs}}$ is the aggregated observation-level visual
representation, and $h(\cdot)$ is an observation-feature aggregation
function used to combine visual representations from different systems. The final representation is given by
\begin{equation}
\tilde c
=
\gamma\,c_{\mathrm{fused}}
+
(1-\gamma)\,c_{\mathrm{obs}},
\end{equation}
where $\gamma$ denotes the causal fusion confidence weight jointly
determined by structural uncertainty and model uncertainty, and $\tilde c$
denotes the final representation used for downstream prediction. \textcolor{black}{The overall
fusion strategy is illustrated in the ACC-CRL model structure shown in
Fig.~\ref{fig:general plan}(b).}

This design introduces an uncertainty-aware observation-supplementation
mechanism. When cross-system structural consistency is high and predictive
uncertainty is low, $\gamma$ becomes large, causing the model to rely primarily
on the causal fusion representation for prediction. Conversely, when structural
mismatch or predictive risk increases, $\gamma$ decreases, prompting the model
to incorporate more fine-grained visual cues from the observation space. This
adaptive supplementation strategy mitigates the risks introduced by unreliable
cross-system causal relations while preserving diagnostically relevant
observation-level information.

Overall, this branch does not replace the CRL process of ACC-CRL, but instead serves as an observation-level complement that enables dynamic balancing between causal abstraction and raw visual information, thereby improving robustness in real-world medical imaging scenarios.

\subsection{Joint Optimization Objective}

Combining the above components, the overall optimization objective of ACC-CRL is
\begin{equation}
\mathcal{L}
=
\mathcal{L}_{\mathrm{sup}}
+
\lambda_{\mathrm{con}}\mathcal{L}_{\mathrm{con}}
+
\lambda_{\mathrm{dis}}\mathcal{L}_{\mathrm{dis}}
+
\lambda_{\mathrm{ali}}\mathcal{L}_{\mathrm{ali}}
+
\lambda_{\mathrm{uaf}}\mathcal{L}_{\mathrm{uaf}},
\label{eq:ACC_crl_total}
\end{equation}
where $\mathcal{L}_{\mathrm{sup}}$ denotes the downstream supervised objective; $\mathcal{L}_{\mathrm{con}}$ learns cross-system shared content representations; $\mathcal{L}_{\mathrm{dis}}$ suppresses mechanism leakage into shared content variables; $\mathcal{L}_{\mathrm{ali}}$ enforces bidirectional response consistency under shared anchors while generating structural mismatch signals; and $\mathcal{L}_{\mathrm{uaf}}$ regularizes sample-level reliability gating in uncertainty-aware fusion (UAF).

Specifically, UAF does not require additional uncertainty annotations, but is jointly driven by structural mismatch signals and task supervision. To maintain consistency between reliability gating and structural consistency, we define
\begin{equation}
\mathcal{L}_{\mathrm{uaf}}
=
\mathbb{E}
\left[
\gamma \cdot e_{\mathrm{str}}
\right].
\label{eq:ACC_crl_luaf}
\end{equation}
This regularizer penalizes assigning large fusion weights under severe structural mismatch, thereby encouraging the model to exploit complementary cross-system information only when structures are reliable, while suppressing cross-system dependencies under structural unreliability or high risk.

From an optimization perspective, $\mathcal{L}_{\mathrm{con}}$ and $\mathcal{L}_{\mathrm{dis}}$ construct shared anchor references, $\mathcal{L}_{\mathrm{ali}}$ exposes cross-system structural mismatch, $\mathcal{L}_{\mathrm{sup}}$ drives task discriminability, and $\mathcal{L}_{\mathrm{uaf}}$ further constrains structural regulation variables. Overall, ACC-CRL realizes the core principle of CGR in latent space: actively constructing and regulating cross-system causal relations through model priors, rather than naively concatenating multi-source information.

\section{Experiments and Results}
\subsection{Experimental Objectives and Methodologies}

The experiments are designed to validate ACC and ACC-CRL from three complementary perspectives: structural mechanism analysis, real-world OOD generalization, and component-level causal attribution.

First, controlled experiments on ColorMNIST are used to analyze the mechanism-level behavior of the proposed framework under explicitly constructed shortcut correlations. 
Rather than focusing solely on classification accuracy, these experiments aim to verify whether ACC-CRL can suppress unstable cross-system dependencies and alter the underlying causal structure in representation space. 
Metrics such as directional bias, counterfactual consistency, retrieval performance, and representation visualization are therefore used as structural evidence for evaluating causal pathway regulation.

Second, real-world multi-center MVI prediction experiments are conducted to evaluate whether the proposed framework generalizes under heterogeneous mechanisms and distribution shifts. 
Compared with synthetic shortcut settings, multi-center medical imaging introduces substantially more complex mechanism variations, including acquisition protocols, scanner differences, and population heterogeneity. 
These experiments primarily validate whether ACC-CRL can maintain transferable shared representations and suppress unstable cross-system correlations under realistic OOD conditions.

Third, ablation studies are designed to analyze the causal roles of different components in ACC-CRL. 
Specifically, mechanism alignment is intended to realize CGT through shared-anchor consistency, while UAF realizes CGR by regulating unreliable cross-system relations. 
The ablations therefore provide component-level evidence connecting the theoretical principles of ACC with their practical realization in ACC-CRL.

\subsection{CSE Analysis Using ColorMNIST}
\begin{table*}[t]
\centering
\footnotesize
\setlength{\tabcolsep}{3.5pt}
\caption{Performance Comparison on ColorMNIST under Different Bias Strengths}
\label{tab:colormnist_results}
\begin{tabular}{c l c c c c c}
\toprule
Bias & Method & OOD Acc & OOD R@1 & ID Acc & CF Same-Class & Dir. ($h$-diff) \\
\midrule
\multirow{3}{*}{0.95}
& Concat. fusion 
& 87.69$\pm$.47 & 92.01$\pm$1.33 & 98.75$\pm$.12 & 96.58$\pm$2.03 & 0.0196$\pm$.0013 \\
& Intervention 
& \textbf{89.39$\pm$.45} & 90.97$\pm$.62 & \textbf{98.98$\pm$.09} & 96.72$\pm$.23 & 0.0184$\pm$.0004 \\
& Intervention + UAF 
& 87.99$\pm$1.18 & \textbf{95.29$\pm$1.12} & 98.87$\pm$.03 & 95.67$\pm$1.10 & \textbf{0.0114$\pm$.0008} \\
\midrule
\multirow{3}{*}{0.98}
& Concat. fusion 
& 78.07$\pm$1.76 & 87.20$\pm$2.32 & \textbf{99.30$\pm$.04} & 94.38$\pm$1.07 & 0.0291$\pm$.0026 \\
& Intervention 
& \textbf{80.36$\pm$.62} & 86.61$\pm$2.10 & 99.16$\pm$.14 & 94.21$\pm$.98 & 0.0272$\pm$.0014 \\
& Intervention + UAF 
& 74.10$\pm$2.04 & \textbf{90.17$\pm$1.02} & 99.18$\pm$.08 & 92.97$\pm$.18 & \textbf{0.0178$\pm$.0006} \\
\midrule
\multirow{3}{*}{0.99}
& Concat. fusion 
& \textbf{69.36$\pm$2.49} & 84.72$\pm$1.93 & \textbf{99.34$\pm$.11} & 92.38$\pm$1.61 & 0.0381$\pm$.0009 \\
& Intervention 
& 67.89$\pm$3.34 & 85.31$\pm$.71 & 99.26$\pm$.30 & 92.36$\pm$2.04 & 0.0388$\pm$.0021 \\
& Intervention + UAF 
& 67.28$\pm$1.81 & \textbf{88.31$\pm$.66} & 99.31$\pm$.07 & 90.02$\pm$1.27 & \textbf{0.0214$\pm$.0009} \\
\bottomrule
\end{tabular}
\end{table*}
\begin{table}[t]
\centering
\small
\caption{Performance Comparison under Different Classifiers}
\begin{tabular}{lcc}
\toprule
Method & Linear Head & k-NN (k=10) \\
\midrule
Concat. fusion  & $\sim$69\% & 86.38\% \\
Intervention & $\sim$67\% & 85.20\% \\
Intervention + UAF & $\sim$67\% & \textbf{89.40\%} \\
\bottomrule
\end{tabular}
\label{tab:knn}
\end{table}

\begin{figure}[!t]
\centering
\includegraphics[width=0.5\textwidth]{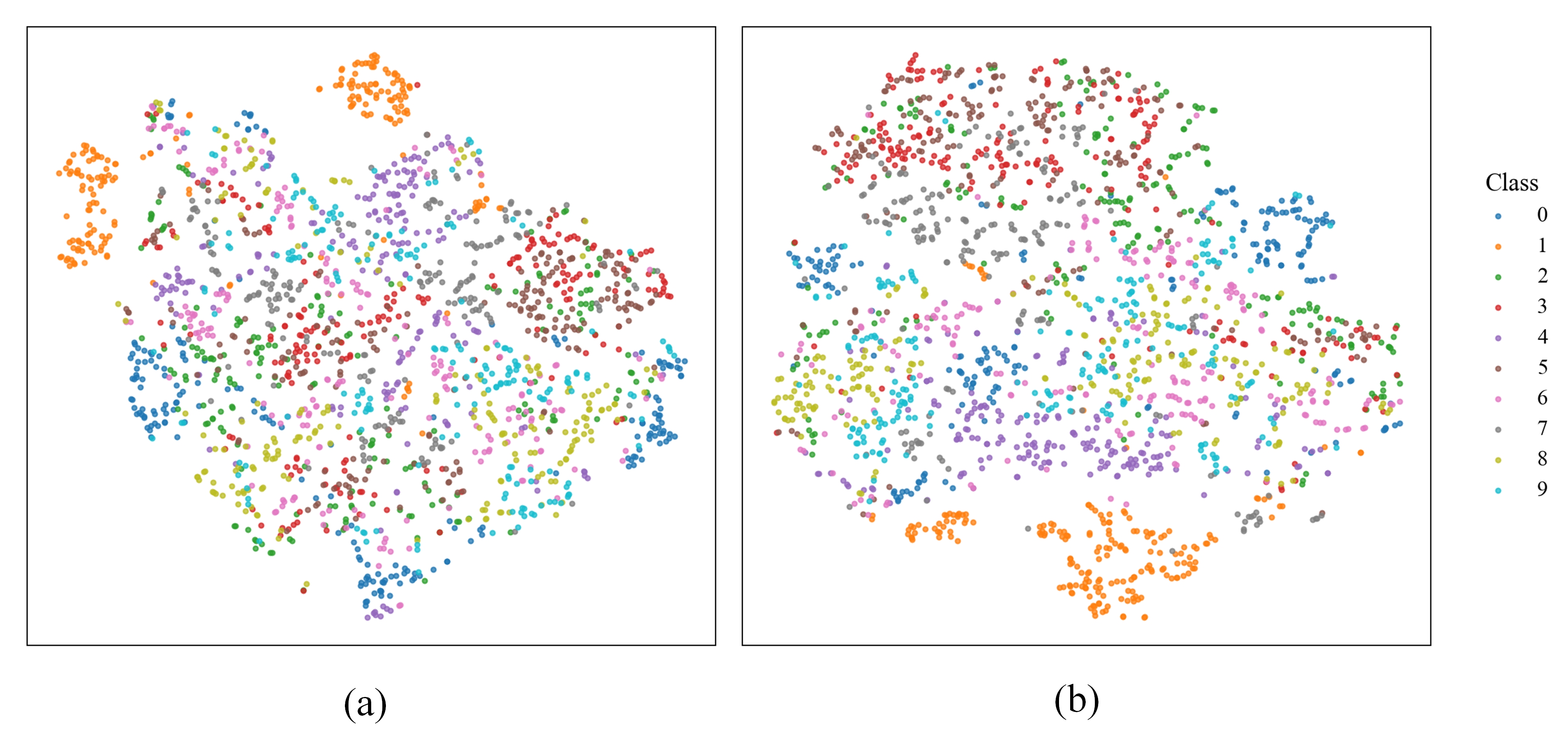}
\caption{t-SNE visualization under high bias (Bias=0.99). Baseline shows severe overlap due to shortcut reliance, while Intervention + UAF yields more separable clusters.}
\label{fig:tsne_colormnist}
\end{figure}

From the perspective of our framework, shortcut causal pathways can be seen as a manifestation of CSE, arising from unstable dependencies between systems. To systematically evaluate the ability of the proposed method to mitigate such spurious causal pathways, we design a controlled experiment using ColorMNIST. By explicitly introducing modality bias, the model is encouraged to learn these unstable correlations during training, offering a clear and interpretable setup for analyzing the dominance of causal pathways. This setup helps assess the effectiveness of CGT in suppressing CSD and the role of CGR in addressing CSE.

We construct a dual-modality dataset:
\begin{itemize}
\item \textbf{Modality A:} Colored digit images, where color is strongly correlated with labels in training, serving as an explicit shortcut signal;
\item \textbf{Modality B:} Grayscale or texture-based representations, preserving structural information and largely independent of color bias.
\end{itemize}

During training, different bias levels (Bias = 0.95, 0.98, 0.99) are used to control color-label correlation. During testing, OOD data (Bias = 0.1) are constructed by breaking this correlation.

From a causal perspective, the evaluation metrics capture key structural properties. Counterfactual consistency (CF Same-Class) measures whether representations capture stable semantics. Directionality ($h$-diff) quantifies directional bias, indicating causal dominance by a single modality. R@1 and k-NN performance reflect local consistency and transferability.

Results reveal fundamental differences across methods. Concatenation-based fusion achieves high ID accuracy but suffers significant OOD degradation with large $h$-diff, indicating reliance on color-induced shortcut pathways. Intervention improves OOD performance but still exhibits directional bias. In contrast, Intervention + UAF significantly reduces $h$-diff and improves R@1, indicating reduced causal dominance and enhanced structural consistency.

Under counterfactual settings, a decrease in CF Same-Class reflects reduced reliance on spurious correlations rather than instability. At high bias levels, OOD accuracy does not significantly improve, highlighting that ACC-CRL prioritizes causal robustness over shortcut-driven discrimination. When replacing the linear classifier with k-NN, Intervention + UAF achieves superior performance, indicating better representation quality.

t-SNE visualizations further confirm that the proposed method produces more compact and separable clusters, demonstrating effective suppression of shortcut pathways.

Overall, the ColorMNIST experiments demonstrate that bias-induced spurious correlations manifest as directional dominance in the representation space, and ACC-CRL effectively mitigates this through mechanism intervention and uncertainty modeling, thereby improving OOD generalization. From a structural perspective, this suggests that the model not only suppresses spurious correlations but also alters the underlying causal structure in the representation space by dynamically regulating cross-modal causal pathways. The reduction in directional bias signifies improved CGR, while the stable semantic clustering reflects the effectiveness of CGT in ensuring consistent causal relationships across systems.
\subsection{MVI Experiments}

Such causal shortcuts induced by mechanism differences are widespread in medical image analysis, and MVI prediction provides a representative example. In this task, the model is expected to leverage complementary information from both intra-tumoral and peri-tumoral regions. However, these two regions are governed by substantially different generative mechanisms, reflecting tumor-internal structural characteristics and changes in the surrounding microenvironment, respectively. Without explicit mechanism modeling, cross-region fusion may easily introduce unstable dependencies driven by region-specific biases, thereby forming non-transferable shortcut causal pathways.

\subsubsection{Patient Recruitment and Inclusion Criteria}

This retrospective study collected data from 386 patients with HCC confirmed by clinical diagnosis and postoperative pathology from two medical institutions: 186 cases from Huzhou Central Hospital and 200 cases from Sichuan Provincial People's Hospital.

The inclusion criteria were as follows:
(1) all cases met the \emph{Clinical Diagnostic Criteria for Primary Liver Cancer} established at the 8th National Academic Conference on Liver Cancer;
(2) dynamic contrast-enhanced MRI was performed within 2 weeks before surgery, with all scans acquired on the same MRI system model (GE Discovery HD750 3.0T), and with complete relevant sequences available; and
(3) laboratory indicators, including AFP, CEA, and CA199, were collected within 1 week before surgery.

The exclusion criteria were as follows:
(1) patients who had received preoperative treatment for intrahepatic lesions, such as radiofrequency ablation (RFA), transarterial chemoembolization (TACE), or radiotherapy;
(2) patients with other types of hepatic tumors; and
(3) patients with inconsistent MRI acquisition parameters or poor image quality that did not satisfy the requirements for imaging--pathology analysis.

After applying the above criteria, 264 patients were finally included in the analysis, including 160 cases from Sichuan Provincial People's Hospital and 104 cases from Huzhou Central Hospital.

\textbf{Cohort 1:} Sichuan Provincial People's Hospital. This cohort included 160 patients collected from November 2019 to January 2021, including 117 MVI-negative cases and 43 MVI-positive cases. To address class imbalance, 43 additional augmented samples were generated for the MVI-positive class. Under the slice-level setting, the 4 slices with the largest ROI were selected for each patient as model inputs, resulting in a total of 812 samples.

\textbf{Cohort 2:} Huzhou Central Hospital. This cohort included 104 patients collected from October 2017 to December 2023, including 78 MVI-negative cases and 26 MVI-positive cases. Similarly, under the slice-level setting, the 4 slices with the largest region-of-interest (ROI) were selected for each patient as model inputs, resulting in a total of 416 samples.

To evaluate the model's generalization ability under distribution shift, Cohort 2 was not merged with Cohort 1, but instead served as an independent external test set for cross-center validation.

\subsubsection{Experimental Setup}

This experimental design enables us to assess whether the proposed method can learn stable pathological representations through causal constraints under substantial cross-center heterogeneity.

The MRI data used in this study consist of two inputs from the arterial phase (ART): the intra-tumoral region and the peri-tumoral region, which are used to validate the effectiveness of the proposed method on a real-world medical task of MVI prediction. Unlike the controlled experiments on ColorMNIST, MVI prediction represents a clinically more complex real-world setting: across different centers, variations in scanning devices, acquisition protocols, patient population distributions, and annotation criteria are common. These factors all lead to changes in modality-specific mechanism variables, making the model more likely to rely on spurious correlations in the training domain and thus suffer significant performance degradation on external data. From the perspective of the proposed multi-causal-system modeling framework, this phenomenon can be understood as follows: during training, the model learns cross-modal shortcut pathways that are statistically correlated within a specific center but unrelated to the intrinsic pathology of the disease, and these pathways become unstable under cross-center testing.

To prevent patient information leakage, all data splits were performed strictly at the patient level. All slices from the same patient were always assigned to the same subset and never appeared simultaneously in the training, validation, and test sets. It should be emphasized that positive-sample augmentation was conducted only after patient-wise splitting and was applied exclusively within the training set; neither the validation set nor the test set contained any augmented samples.

During model training and evaluation, we adopted a slice-level setting: for each patient, the 4 slices with the largest ROI were selected as input samples, and both training and testing were performed at the slice level. Therefore, the AUC, ACC, BACC, Specificity, and F1 reported in this paper are all slice-level metrics rather than patient-level aggregated results. In particular, BACC is additionally reported to provide a more reliable evaluation under OOD settings with potentially imbalanced sample distributions across classes.

In addition, the internal and external cohorts followed the same ROI selection rules and preprocessing pipeline. Specifically, for both cohorts, the 4 slices with the largest ROI were selected for each patient according to the same criterion, and identical cropping, normalization, and size standardization procedures were applied. The external cohort was used only for out-of-distribution testing and did not participate in model training, validation, hyperparameter selection, or data augmentation.

\subsubsection{Experimental Results}

\begin{table*}[htbp]
\centering
\small
\caption{Performance Comparison on the Internal ODD-Data Cohort. Ours Denotes the Final ACC-CRL Model with Evidential Fusion.}
\label{tab:internal_ok3d1}
\begin{tabular}{lccccc}
\toprule
Method & AUC & ACC & BACC & Specificity & F1 \\
\midrule

MultiOOD~\cite{dong2024multiood} & \textbf{0.9503} $\pm$ 0.0225 & 0.8056 $\pm$ 0.0507 & 0.8182 $\pm$ 0.0504 & 0.7375 $\pm$ 0.1611 & 0.7966 $\pm$ 0.0270 \\
SupContrast~\cite{khosla2020supervised} & 0.8974 $\pm$ 0.0239 & 0.7961 $\pm$ 0.0303 & 0.8093 $\pm$ 0.0187 & 0.7684 $\pm$ 0.1048 & 0.7722 $\pm$ 0.0545 \\
IRM~\cite{arjovsky2019invariant} & 0.6834 $\pm$ 0.0995 & 0.5728 $\pm$ 0.1184 & 0.6073 $\pm$ 0.0660 & 0.4101 $\pm$ 0.2129 & 0.6058 $\pm$ 0.1185 \\
Mixup~\cite{zhang2017mixup}  & 0.7680 $\pm$ 0.0350 & 0.6788 $\pm$ 0.0279 & 0.6885 $\pm$ 0.0242 & 0.6825 $\pm$ 0.0878 & 0.6391 $\pm$ 0.0611 \\
Group DRO~\cite{sagawa2019distributionally}  & 0.7801 $\pm$ 0.0253 & 0.6935 $\pm$ 0.0471 & 0.6967 $\pm$ 0.0222 & 0.6629 $\pm$ 0.1065 & 0.6620 $\pm$ 0.0682 \\
VREx~\cite{krueger2021out}  & 0.9025 $\pm$ 0.0340 & \textbf{0.8459} $\pm$ 0.0282 & \textbf{0.8355} $\pm$ 0.0369 & \textbf{0.8561} $\pm$ 0.0760 & \textbf{0.8082} $\pm$ 0.0625 \\
CDANN~\cite{li2018deep}  & 0.8918 $\pm$ 0.0569 & 0.8109 $\pm$ 0.0650 & 0.8299 $\pm$ 0.0478 & 0.7821 $\pm$ 0.1210 & 0.7933 $\pm$ 0.0724 \\
ACC-CRL & 0.9137 $\pm$ 0.0368 & 0.7842 $\pm$ 0.0775 & 0.8094 $\pm$ 0.0576 & 0.7374 $\pm$ 0.1237 & 0.7725 $\pm$ 0.0807  \\
\bottomrule
\end{tabular}
\label{tab:iid}
\end{table*}

\begin{table*}[htbp]
\centering
\small
\caption{Performance Comparison on the External HZ Cohort. Ours Denotes the Final ACC-CRL Model with Evidential Fusion.}
\label{tab:external_hz}
\begin{tabular}{lccccc}
\toprule
Method & AUC & ACC & BACC & Specificity & F1 \\
\midrule

MultiOOD~\cite{dong2024multiood}  & 0.5211 $\pm$ 0.0782 & 0.6294 $\pm$ 0.0635 & 0.5236 $\pm$ 0.0446 & 0.7395 $\pm$ 0.1634 & 0.2631 $\pm$ 0.1086 \\
SupContrast~\cite{khosla2020supervised} & 0.5442 $\pm$ 0.0370 & 0.6784 $\pm$ 0.0555 & 0.5533 $\pm$ 0.0231 & 0.8086 $\pm$ 0.1173 & 0.3036 $\pm$ 0.0776 \\
IRM~\cite{arjovsky2019invariant} & \textbf{0.7143} $\pm$ 0.0048 & 0.5441 $\pm$ 0.1312 & 0.6251 $\pm$ 0.0465 & 0.4599 $\pm$ 0.2250 & \textbf{0.4738} $\pm$ 0.0275 \\
Mixup~\cite{zhang2017mixup}  & 0.5709 $\pm$ 0.0638 & 0.6402 $\pm$ 0.0621 & 0.5587 $\pm$ 0.0356 & 0.7250 $\pm$ 0.1164 & 0.3508 $\pm$ 0.0597 \\
Group DRO~\cite{sagawa2019distributionally}  & 0.5511 $\pm$ 0.0736 & 0.6255 $\pm$ 0.0602 & 0.5463 $\pm$ 0.0574 & 0.7079 $\pm$ 0.1161 & 0.3282 $\pm$ 0.0908 \\
VREx~\cite{krueger2021out}  & 0.5785 $\pm$ 0.0363 & 0.7132 $\pm$ 0.0230 & 0.5248 $\pm$ 0.0275 & \textbf{0.9092} $\pm$ 0.0514 & 0.1838 $\pm$ 0.1087 \\
CDANN~\cite{li2018deep} & 0.5872 $\pm$ 0.0356 & 0.7078 $\pm$ 0.0250 & 0.5370 $\pm$ 0.0406 & 0.8855 $\pm$ 0.0595 & 0.2247 $\pm$ 0.1338 \\
ACC-CRL& 0.6840 $\pm$ 0.0296 & \textbf{0.7240} $\pm$ 0.0321 & \textbf{0.6345} $\pm$ 0.0267 & 0.8171 $\pm$ 0.0615 & 0.4522 $\pm$ 0.0437 \\ 
\bottomrule
\end{tabular}
\end{table*}

\begin{figure}[t]
\centering
\includegraphics[width=0.5\textwidth]{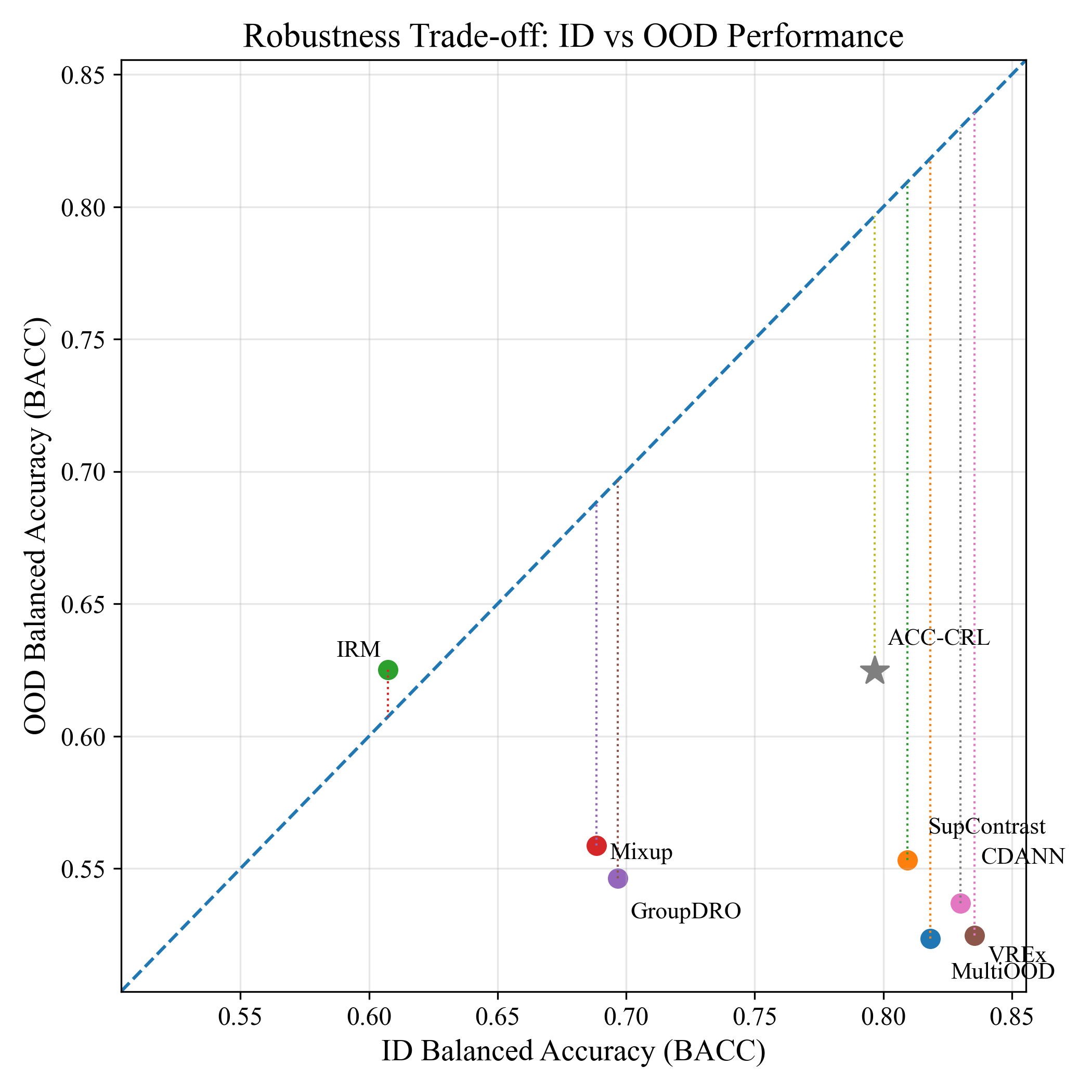}
\caption{
Robustness trade-off between ID and OOD performance measured by BACC.
Each point denotes a method, and the diagonal line indicates equal performance across environments.
Methods located closer to the diagonal exhibit smaller generalization gaps.
ACC-CRL lies in a more favorable region, maintaining competitive ID performance while achieving stronger OOD generalization,
indicating improved robustness under distribution shift.
}
\label{fig:tradeoff}
\end{figure}

The experimental results are summarized in Tables~\ref{tab:iid} and~\ref{tab:external_ood}, which report the performance of different methods under ID and OOD settings, respectively.

From the results, it is evident that different methods exhibit significant discrepancies across distributions. Some approaches (e.g., MultiOOD) achieve strong performance on ID data but suffer severe degradation under OOD conditions (e.g., AUC drops from 0.9503 to 0.5211), indicating that they rely heavily on statistical correlations in the training distribution and consequently learn non-transferable shortcut pathways.

In contrast, ACC-CRL maintains competitive performance on ID data while achieving consistently stronger performance under OOD settings compared to most baselines. Although it does not always achieve the best ID scores (e.g., compared to MultiOOD or VREx), it exhibits substantially smaller performance degradation across distributions, demonstrating improved cross-distribution generalization. This suggests that the model tends to capture more invariant and transferable features rather than relying on domain-specific biased patterns.

Furthermore, Fig.~\ref{fig:tradeoff} provides a visual summary of the ID--OOD trade-off. Most baseline methods are located in the ``high-ID, low-OOD'' region, indicating substantial generalization gaps. In comparison, ACC-CRL lies closer to the diagonal, achieving a better balance by maintaining competitive ID performance while exhibiting reduced degradation under distribution shift.

In addition, from the perspective of BACC and F1 metrics, ACC-CRL demonstrates more balanced performance across distributions compared to several baselines. While it may not always achieve the highest F1 score (e.g., IRM under OOD), it avoids extreme bias toward specific classes, indicating more stable and reliable predictions under distribution shift. This property is particularly important in medical applications, where consistent performance across varying data distributions is essential.

Overall, the results indicate that ACC-CRL improves OOD generalization while maintaining competitive ID performance, achieving a more favorable trade-off between accuracy and robustness. This result indicates that ACC-CRL does not merely enhance robustness, but dynamically adjusts cross-region causal relationships under cross-center mechanism heterogeneity, thereby enabling adaptive construction at the level of causal structure.

\subsubsection{MVI Ablation Study}
\begin{table*}[t]
\centering
\footnotesize
\setlength{\tabcolsep}{4pt}
\caption{Performance comparison on the internal ID (in-distribution) cohort. 
Ours denotes the full ACC-CRL model integrating mechanism alignment and UAF.}
\label{tab:internal_ok3d}
\begin{tabular}{p{4.0cm}ccccc}
\toprule
Method & AUC & ACC & BACC & Specificity & F1 \\
\midrule

Baseline (no alignment, no UAF) 
& 0.8810 $\pm$ 0.0350 
& 0.7452 $\pm$ 0.0552 
& 0.7746 $\pm$ 0.0411 
& 0.6151 $\pm$ 0.0803 
& 0.7517 $\pm$ 0.0633 \\

Alignment only 
& 0.8854 $\pm$ 0.0392 
& 0.7626 $\pm$ 0.0648 
& 0.7957 $\pm$ 0.0396 
& 0.6806 $\pm$ 0.1275 
& 0.7601 $\pm$ 0.0712 \\

UAF (structure only)
& 0.8907 $\pm$ 0.0476
& 0.7517 $\pm$ 0.1061
& 0.7776 $\pm$ 0.0785
& 0.6395 $\pm$ 0.1702
& 0.7583 $\pm$ 0.0964
 \\

UAF (model only)
& 0.8926 $\pm$ 0.0600
& 0.7438 $\pm$ 0.1511
& 0.7799 $\pm$ 0.1055
& 0.6855 $\pm$ 0.2567
& 0.7492 $\pm$ 0.1170\\

\textbf{ACC-CRL (alignment + UAF)}
& \textbf{0.9137 $\pm$ 0.0368} 
& \textbf{0.7842 $\pm$ 0.0775} 
& \textbf{0.8094 $\pm$ 0.0576} 
& \textbf{0.7374 $\pm$ 0.1237 }
& \textbf{0.7725 $\pm$ 0.0807} \\

\bottomrule
\end{tabular}
\end{table*}

\begin{table*}[t]
\centering
\footnotesize
\setlength{\tabcolsep}{4pt}
\caption{Performance comparison on the external HZ cohort (out-of-distribution, OOD), evaluating cross-center generalization. 
Ours denotes the full ACC-CRL model integrating mechanism alignment and UAF.}
\label{tab:external_ood}
\begin{tabular}{p{4.0cm}ccccc}
\toprule
Method & AUC & ACC & BACC & Specificity & F1 \\
\midrule

Baseline (no alignment, no UAF) 
& 0.6558 $\pm$ 0.0340 
& 0.6637 $\pm$ 0.0563 
& 0.6257 $\pm$ 0.0298 
& 0.7033 $\pm$ 0.0975 
& \textbf{0.4541 $\pm$ 0.0323} \\

Alignment only 
& 0.6596 $\pm$ 0.0437 
& 0.6912 $\pm$ 0.0280 
& 0.6169 $\pm$ 0.0364 
& 0.7684 $\pm$ 0.0576 
& 0.4296 $\pm$ 0.0594 \\

UAF (structure only)
& 0.6699 $\pm$ 0.0299
& 0.6578 $\pm$ 0.0589
& 0.6243 $\pm$ 0.0178
& 0.6928 $\pm$ 0.1207
& 0.4500 $\pm$ 0.0268
 \\

UAF (model only)
& 0.6625 $\pm$ 0.0430
& 0.6887 $\pm$ 0.0765
& 0.6184 $\pm$ 0.0268
& 0.7618 $\pm$ 0.1564
& 0.4298 $\pm$ 0.0486\\

\textbf{ACC-CRL (alignment + UAF)} 
& \textbf{0.6840 $\pm$ 0.0296} 
& \textbf{0.7240 $\pm$ 0.0321} 
& \textbf{0.6345 $\pm$ 0.0267} 
& \textbf{0.8171 $\pm$ 0.0615} 
& 0.4522 $\pm$ 0.0437 \\

\bottomrule
\end{tabular}
\end{table*}
The ablation results demonstrate that different components exhibit complementary effects under both ID and OOD settings. Mechanism alignment improves CGT, while UAF enhances CGR by suppressing CSE.

On the ID, introducing mechanism alignment alone leads to consistent improvements over the baseline across AUC, ACC, and BACC. This suggests that cross-modal mechanism alignment effectively mitigates disturbances caused by heterogeneous generative mechanisms, encouraging the model to rely more on shared causal content rather than spurious local correlations.

When incorporating UAF, the performance exhibits moderate fluctuations. While UAF enhances AUC in certain configurations (e.g., structure-only and model-only variants), it does not consistently improve ACC and BACC compared to alignment alone. This indicates that, under in-distribution settings, uncertainty modeling may suppress some predictive but potentially unstable correlations, resulting in a trade-off between peak performance and robustness.

On the OOD, the role of each component becomes more evident. Mechanism alignment alone improves AUC and ACC compared to the baseline, but leads to decreased BACC, suggesting limited ability to handle distribution shifts across modalities. In contrast, integrating UAF leads to more balanced performance, improving AUC and BACC relative to both the baseline and alignment-only variants, while maintaining competitive F1 scores. This indicates that uncertainty modeling helps identify and down-weight unreliable cross-modal interactions under distribution shift.

Overall, mechanism alignment enhances CGT at the representation level, while uncertainty modeling regulates the reliability of multi-modal fusion. Their combination enables ACC-CRL to achieve the best overall performance, yielding consistent gains on ID data and more robust, balanced generalization under OOD settings.

\subsection{Mechanism Disentanglement Analysis}

\begin{figure}[!t]
    \centering
    \includegraphics[width=0.5\textwidth]{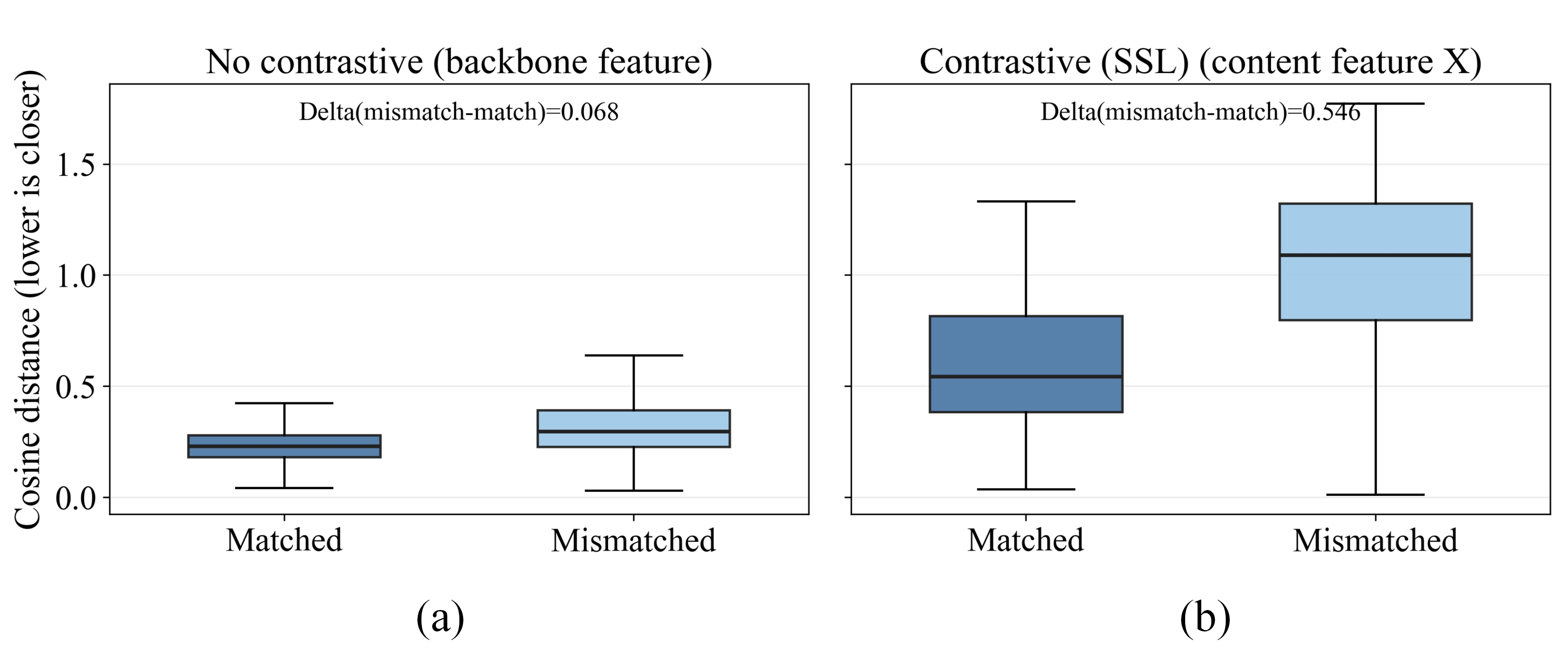}
\caption{Comparison of cosine distance distributions between matched and mismatched sample pairs before and after contrastive learning. (a) Without contrastive learning, the distance distributions largely overlap, with a small gap ($\Delta = 0.068$), indicating weak semantic structure. (b) After applying contrastive learning, although overall distances increase, the gap between mismatched and matched pairs becomes significantly larger ($\Delta = 0.546$), suggesting enhanced semantic separation in the representation space.}
    \label{fig:SSL}
\end{figure}

To further analyze the effect of contrastive learning on the representation space, we compare the cosine distance distributions between matched and mismatched sample pairs, as shown in Fig.~\ref{fig:SSL}. Here, matched pairs refer to semantically consistent samples, while mismatched pairs correspond to semantically inconsistent ones.

Without contrastive learning (Fig.~\ref{fig:SSL}(a)), the distance distributions of matched and mismatched pairs largely overlap, with a small difference ($\Delta = \text{mismatched} - \text{matched} = 0.068$), indicating that the model fails to learn a discriminative structure and that semantic consistency is not well captured in the representation space.

In contrast, after introducing contrastive learning (Fig.~\ref{fig:SSL}(b)), although the overall distance distribution expands (i.e., both matched and mismatched distances increase), the relative gap between them is significantly enlarged ($\Delta = 0.546$). This suggests that contrastive learning does not simply compress the feature space but instead increases the separation between semantically different samples while maintaining relative proximity among semantically consistent ones, thereby forming a clearer discriminative structure.

Therefore, compared to absolute distances, the \emph{gap between matched and mismatched pairs} better reflects the learned semantic structure. This result verifies that contrastive learning effectively enhances semantic consistency in representations, providing a more stable foundation for subsequent mechanism alignment and causal structure modeling.

\begin{figure}[!t]
    \centering
    \includegraphics[width=0.5\textwidth]{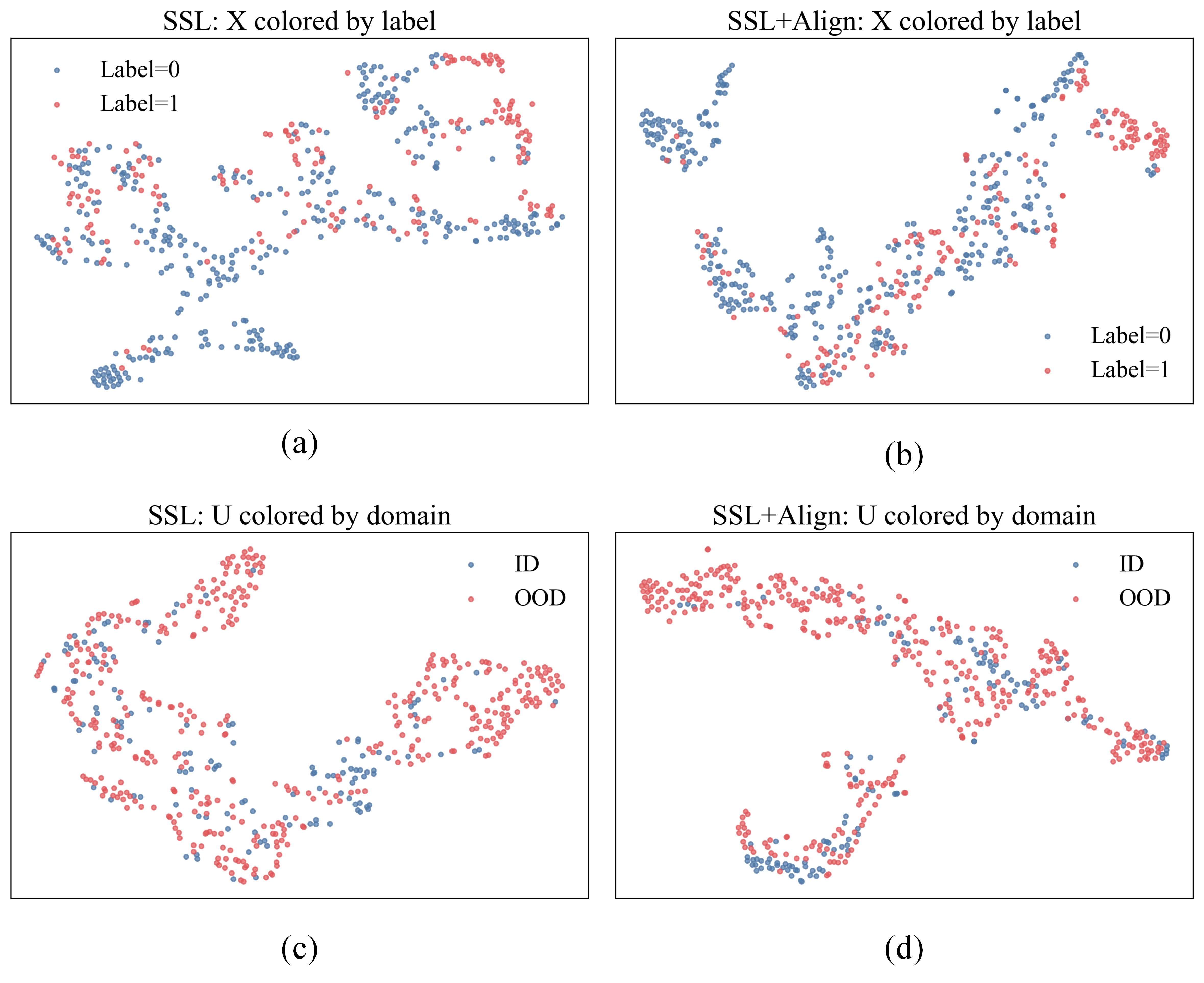}
\caption{t-SNE visualization of latent representations before and after mechanism alignment. (a)(c) show representations learned by contrastive learning (SSL), while (b)(d) show results after introducing mechanism alignment (SSL+Align). The top row corresponds to shared representation \(Z\) (colored by class), and the bottom row corresponds to mechanism representation \(U\) (colored by domain, where ID denotes training distribution and OOD denotes out-of-distribution data). After alignment, \(Z\) becomes more compact and separable across classes, while \(U\) exhibits clearer domain separation, indicating improved disentanglement between shared semantics and mechanism-specific factors.}
    \label{fig:tnes}
\end{figure}

To further investigate the impact of mechanism alignment on latent representations, we visualize the shared representation \(X\) and mechanism-specific representation \(U\) using t-SNE, as shown in Fig.~\ref{fig:tnes}.

During the contrastive learning stage (SSL) (Fig.~\ref{fig:tnes}(a), (c)), the shared representation \(X\) partially captures class information but still exhibits noticeable overlap across categories, indicating limited discriminative power. Meanwhile, the mechanism representation \(U\) does not show clear separation across domains (ID vs. OOD), suggesting that the model has not yet effectively captured mechanism variations associated with distribution shifts.

After introducing mechanism alignment (SSL+Align) (Fig.~\ref{fig:tnes}(b), (d)), the representation structure undergoes significant changes. First, \(X\) forms more compact and well-separated clusters along class dimensions, indicating enhanced shared semantic representation relevant to the task. Second, \(U\) shows clearer separation across domains, suggesting that the model successfully encodes distribution-specific mechanism information into an independent representation space.

These observations indicate that mechanism alignment not only improves discriminative capability but also promotes functional disentanglement between shared semantics and mechanism factors, thereby facilitating the learning of more stable representations with improved cross-distribution generalization.

\subsection{Uncertainty Modeling Analysis}

\begin{figure*}[!t]
    \centering
    \includegraphics[width=1\textwidth]{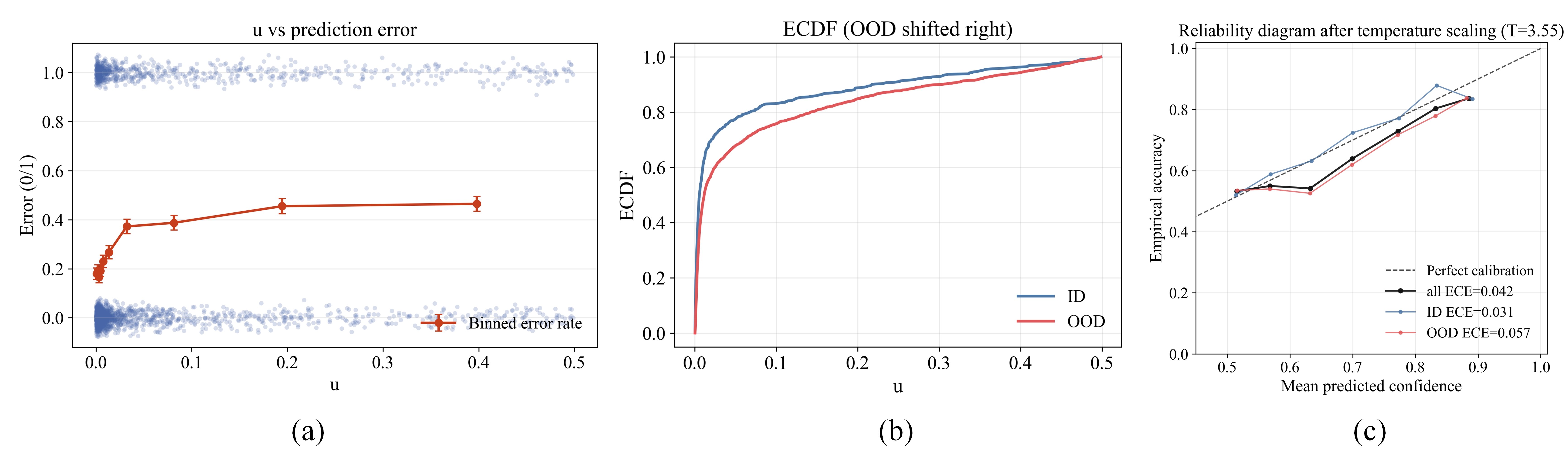}
\caption{Uncertainty modeling analysis. (a) Relationship between predictive uncertainty $u=1-\max(p)$ and classification error, where $p$ denotes predicted class probabilities. Results show a positive correlation between uncertainty and prediction risk. (b) Empirical cumulative distribution functions (ECDF) of uncertainty across domains, where ID denotes in-distribution data and OOD denotes out-of-distribution data. The rightward shift of OOD indicates higher uncertainty under distribution shift. (c) Reliability diagram after temperature scaling, evaluating probability calibration. ECE (expected calibration error) measures the gap between confidence and accuracy, showing good calibration performance.}
    \label{fig:uncertainty_analysis}
\end{figure*}

To evaluate the effectiveness of the proposed uncertainty module, we analyze predictive uncertainty from three perspectives: error correlation, cross-domain sensitivity, and probability calibration, as shown in Fig.~\ref{fig:uncertainty_analysis}.

First, from the perspective of error correlation (Fig.~\ref{fig:uncertainty_analysis}(a)), predictive uncertainty \(u\), defined as the complement of the fusion confidence weight, i.e., \(u=1-\gamma\), exhibits a clear positive correlation with classification error. As \(u\) increases, the binned error rate consistently rises, indicating that samples with lower fusion reliability are associated with greater prediction risk. This demonstrates that the proposed uncertainty measure effectively captures the reliability of model predictions rather than serving as a heuristic score.

Second, regarding cross-domain sensitivity (Fig.~\ref{fig:uncertainty_analysis}(b)), the uncertainty distribution for the external domain (OOD) is shifted to the right compared to the internal domain (ID), meaning that OOD samples tend to have higher uncertainty. This indicates that the model can detect structural shifts under distribution changes and respond accordingly by increasing uncertainty. This observation is consistent with the design motivation of using structural uncertainty to regulate cross-modal fusion strength.

Finally, from the perspective of probability calibration (Fig.~\ref{fig:uncertainty_analysis}(c)), after applying temperature scaling, the model demonstrates good calibration performance across both overall and domain-specific data. Specifically, the overall ECE is 0.042, while the internal and external domains achieve 0.031 and 0.057, respectively. This indicates strong alignment between predicted confidence and empirical accuracy. Good calibration further suggests that both the predicted confidence and derived uncertainty are reliable.

Overall, these results demonstrate that the proposed uncertainty modeling not only effectively captures prediction risk but also responds to cross-domain distribution shifts while maintaining good probabilistic consistency after calibration. These findings validate the reliability of the uncertainty module and provide empirical support for subsequent uncertainty-aware adaptive fusion.

\section{Conclusion}

In this paper, we investigate the generalization problem in fusion learning under OOD settings from a causal perspective. We highlight that different input sources correspond to heterogeneous generative mechanisms, and cross-center distribution shifts can be fundamentally understood as variations in causal mechanisms. Under such conditions, conventional fusion methods often introduce unstable shortcut pathways driven by statistical correlations, leading to performance degradation, particularly due to CSE and CSD.

To address these issues, we propose the ACC framework, which formulates multi-source fusion as a problem of structural reconfiguration across multiple causal systems. The framework establishes CGT \textit{via} shared causal anchors to mitigate CSD, models structural changes through CGR, and regulates the fusion process via uncertainty modeling to suppress CSE.

Building on this framework, we introduce ACC-CRL as a learnable implementation. The method learns stable causal content representations through content-mechanism disentanglement and interventional consistency alignment, and integrates UAF to adaptively suppress unreliable cross-system pathways, thereby improving robustness against CSE and CSD.

Experiments on both synthetic data (ColorMNIST) and real-world multi-center medical imaging tasks (MVI prediction) demonstrate that the proposed method significantly improves OOD generalization while maintaining ID performance, validating the effectiveness of causal modeling based on CGT, CGR, and uncertainty modeling.

Future work will explore more complex multi-causal-system modeling under diverse input conditions and incorporate more advanced structure learning techniques to further enhance interpretability and clinical applicability, ultimately refining the adaptation of CGT and CGR to address CSD and CSE in real-world scenarios.

\bibliographystyle{IEEEtran}
\bibliography{references}

\appendix

\subsection{First-Order Interpretation of Response Alignment}
\label{appendix_a}

We provide a first-order interpretation of the response alignment objective used in ACC-CRL. 
For each system $m\in\{1,2\}$, let $X^{(m)}$ denote the shared causal content representation and let
\[
R^{(m)} = f_Y^{(m)}(X^{(m)},U^{(m)})
\]
be the corresponding system response, where $U^{(m)}$ captures system-specific mechanism information.

Consider a small intervention on the shared causal content,
\[
do(X^{(m)} \leftarrow X^{(m)}+\delta),
\qquad
\delta \sim \mathcal N(0,I), \quad \|\delta\|\rightarrow 0 .
\]
The induced response variation of system $m$ is defined as
\[
\Delta R^{(m)}\left(X^{(m)},\delta\right)
=
f_Y^{(m)}\left(X^{(m)}+\delta,U^{(m)}\right)
-
f_Y^{(m)}\left(X^{(m)},U^{(m)}\right).
\]

Under the shared-anchor assumption, response consistency requires that two systems exhibit consistent local response changes under the same content intervention. This motivates the following interventional response alignment objective:
\[
\mathcal L_{\mathrm{ali}}^{\mathrm{int}}
=
\mathbb E_{X,\delta}
\left[
\left\|
\Delta R^{(1)}(X^{(1)},\delta)
-
\Delta R^{(2)}(X^{(2)},\delta)
\right\|_2^2
\right].
\]

\subsubsection{From Response Consistency to Jacobian Consistency}

Assume that $f_Y^{(1)}$ and $f_Y^{(2)}$ are continuously differentiable with respect to the shared content variable. 
By first-order Taylor expansion, we have
\[
\Delta R^{(m)}\left(X^{(m)},\delta\right)
=
J_m\left(X^{(m)}\right)\delta
+
\mathcal O\left(\|\delta\|^2\right),
\]
where
\[
J_m\left(X^{(m)}\right)
=
\frac{\partial f_Y^{(m)}\left(X^{(m)},U^{(m)}\right)}{\partial X^{(m)}}
\]
denotes the Jacobian of the system response with respect to the shared content representation.

Substituting the first-order approximation into $\mathcal L_{\mathrm{ali}}^{\mathrm{int}}$ gives
\[
\mathcal L_{\mathrm{ali}}^{\mathrm{int}}
\approx
\mathbb E_{\delta}
\left[
\left\|
\left(
J_1\left(X^{(1)}\right) - J_2\left(X^{(2)}\right)
\right)\delta
\right\|_2^2
\right].
\]
Let
\[
A = J_1\left(X^{(1)}\right) - J_2\left(X^{(2)}\right).
\]
Since $\delta$ follows an isotropic distribution with covariance identity matrix,
\[
\mathbb E_{\delta}\left[\|A\delta\|_2^2\right]
=
\|A\|_F^2 .
\]
Therefore,
\[
\mathcal L_{\mathrm{ali}}^{\mathrm{int}}
\propto
\left\|
J_1\left(X^{(1)}\right) - J_2\left(X^{(2)}\right)
\right\|_F^2 .
\]

This shows that, in the small-intervention limit, enforcing response consistency under shared anchors is asymptotically equivalent to enforcing first-order mechanism consistency between the two systems.

\subsection{Theoretical Extension: Suppressing Spurious Cross-System Correlations}
\label{sec:spurious_correlation}

We further discuss why the proposed response alignment can help suppress spurious correlations shared across heterogeneous systems. 
Existing multi-modal contrastive learning analyses often assume that modality-specific style factors are independent across modalities. 
However, this assumption may not hold in real-world medical imaging, where scanning protocols, center-specific acquisition patterns, or annotation biases may induce correlated non-causal mechanisms across different input sources.

Let the observations from two causal systems be
\begin{align}
T^{(m)} = g^{(m)}\left(X, U^{(m)}\right), \qquad m\in\{1,2\},
\end{align}
where \(X\) denotes the shared causal content variable and \(U^{(m)}\) denotes the system-specific mechanism factor. 
When
\begin{align}
U^{(1)} \not\!\perp U^{(2)},
\end{align}
the two systems may share spurious mechanism-dependent components. 
As a result, representation learning based only on observational consistency may align not only the stable content \(X\), but also correlated non-causal factors induced by \(U^{(1)}\) and \(U^{(2)}\).

In ACC-CRL, each encoder decomposes the observation as
\begin{align}
\left(X^{(m)}, U^{(m)}\right)=C_m\left(T^{(m)}\right),
\end{align}
where \(X^{(m)}\) is encouraged to capture shared causal content, while \(U^{(m)}\) preserves system-specific mechanism information. 
However, contrastive consistency alone may still be insufficient when spurious mechanism factors are correlated across systems.

\subsubsection{Interventional Suppression of Spurious Mechanism Correlations}
\label{prop:spurious}
Assume that the causal content variable \(X\) is independent of the mechanism variables \(U^{(m)}\), and that spurious correlations between \(U^{(1)}\) and \(U^{(2)}\) are not causally involved in the stable downstream response. 
Then enforcing response alignment under shared anchors encourages the learned content representations \(X^{(1)}\) and \(X^{(2)}\) to preserve intervention-sensitive causal content while reducing reliance on spurious mechanism-correlated components.

\begin{proof}
Consider the causal structure
\[
X \rightarrow T^{(m)}, \qquad U^{(m)} \rightarrow T^{(m)}, \qquad m\in\{1,2\}.
\]
If \(U^{(1)}\) and \(U^{(2)}\) are spuriously correlated, then the joint observations are affected by the non-causal path
\[
T^{(1)}
\leftarrow
U^{(1)}
\leftrightarrow
U^{(2)}
\rightarrow
T^{(2)}.
\]
Therefore, purely observational alignment may capture shared information along this path, leading to content representations contaminated by correlated mechanism-dependent factors.

ACC-CRL introduces bidirectional response alignment:
\begin{align}
\mathcal L_{\mathrm{ali}}
=
\sum_i
\left(
\left\|
\hat R_i^{(2\leftarrow 1)} - R_i^{(2)}
\right\|_2^2
+
\left\|
\hat R_i^{(1\leftarrow 2)} - R_i^{(1)}
\right\|_2^2
\right),
\end{align}
where \(\hat R^{(2\leftarrow 1)}\) and \(\hat R^{(1\leftarrow 2)}\) are cross-system responses reconstructed under shared-anchor representations.

Now consider a small intervention on the shared content representation,
\[
do(X^{(m)} \leftarrow X^{(m)}+\delta),
\qquad
\delta \perp U^{(1)},U^{(2)}.
\]
Since the perturbation acts on the content variable rather than on the mechanism variables, only components that are stably associated with the causal content can induce consistent cross-system response changes. 
In contrast, components caused by the spurious path \(U^{(1)}\leftrightarrow U^{(2)}\) are not intervention-sensitive with respect to \(X^{(m)}\), and their contribution cannot reliably explain the response variation under content interventions.

Thus, minimizing \(\mathcal L_{\mathrm{ali}}\) encourages the encoders to preserve content components that support stable cross-system response consistency, while reducing dependence on correlated mechanism-specific components. 
This weakens the influence of the spurious path induced by \(U^{(1)}\leftrightarrow U^{(2)}\), yielding shared representations that are more closely related to causal content.
\end{proof}

When the test-time mechanism distribution \(P(U^{(1)},U^{(2)})\) changes, representations relying on spurious mechanism correlations are likely to become unstable. 
By encouraging response consistency under shared anchors, ACC-CRL reduces such reliance and improves robustness under mechanism shifts. 
This provides a theoretical explanation for the improved OOD behavior observed in the Mechanism-OOD and ColorMNIST experiments.

\subsection{From Structural Credibility to Adaptive Gating}
\label{subsec:gamma_derivation}

In the ACC framework, each candidate cross-system edge \(e\in\Delta E\) is associated with a structural confidence variable \(\pi_e\), and its credibility is measured by the expected confidence
\begin{equation}
\tilde C_{\mathrm{cross}}(e)
=
\mathbb E[C(e)\mid \pi_e^{\mathrm{cross}}].
\end{equation}
In ACC-CRL, this abstract graph-level credibility is instantiated as a differentiable sample-wise gating variable.

Specifically, we view the reliability of a cross-system relation as a latent Bernoulli variable
\begin{equation}
C(e)\in\{0,1\},
\end{equation}
where \(C(e)=1\) indicates that the relation is reliable, and \(C(e)=0\) indicates that it should be weakened or suppressed. 
Since the true edge reliability is unobserved, ACC-CRL estimates it using two uncertainty sources: the structural mismatch \(e_{\mathrm{str}}\) induced by bidirectional response alignment and the predictive uncertainty \(u_{\mathrm{unc}}\).

The structural uncertainty is defined as
\begin{equation}
u_{\mathrm{str}}
=
\sigma\left(\frac{e_{\mathrm{str}}}{\tau}\right),
\end{equation}
and the adaptive gate is defined as
\begin{equation}
\gamma
=
1-
\frac{1}{2}
\left(
u_{\mathrm{str}} + u_{\mathrm{unc}}
\right).
\end{equation}

This gate can be interpreted as a differentiable approximation to the posterior reliability of a candidate cross-system relation:
\begin{equation}
\gamma
\approx
P(C(e)=1\mid e_{\mathrm{str}}, u_{\mathrm{unc}}).
\end{equation}
Therefore, its expected credibility can be approximated as
\begin{equation}
\begin{aligned}
\tilde C_{\mathrm{cross}}(e)
&=
\mathbb E[C(e)\mid e_{\mathrm{str}}, u_{\mathrm{unc}}] \\
&\approx
1\cdot \gamma + 0\cdot (1-\gamma) \\
&=
\gamma .
\end{aligned}
\end{equation}

The fusion representation is then computed as
\begin{equation}
c_{\mathrm{fused}}
=
\gamma c_{\mathrm{base}}
+
(1-\gamma)X^{(1)}.
\end{equation}
Thus, when the estimated cross-system relation is reliable, ACC-CRL strengthens dual-system fusion through \(c_{\mathrm{base}}\); otherwise, it suppresses uncertain cross-system information and falls back to the conservative reference representation \(X^{(1)}\).

This provides a probabilistic interpretation of the adaptive gate \(\gamma\): it serves as a sample-wise neural approximation of cross-system edge credibility, bridging the graph-level structural regulation in ACC and the differentiable fusion mechanism in ACC-CRL.

\end{document}